\documentclass[preprint,12pt]{elsarticle}
\usepackage{natbib}
\usepackage{hyperref}
\usepackage{caption}
\usepackage{amssymb}
\usepackage{amsmath}
\usepackage{booktabs}
\usepackage{xcolor}
\usepackage{verbatim}
\usepackage{listings}

\makeatletter
\def\ps@pprintTitle{%
  \let\@oddhead\@empty
  \let\@evenhead\@empty
  \def\@oddfoot{\footnotesize
    arXiv preprint — June 2025\hfill}%
  \let\@evenfoot\@oddfoot}
\makeatother

\begin{document}

\begin{frontmatter}



\title{Graph-Based Deep Learning for Component Segmentation of Maize Plants }



\author[label1]{\uppercase{J.\,I.\ Ruiz}}
\ead{jesus.ruiz@cinvestav.com}
\author[label1]{\uppercase{A.\ Mendez-Vazquez}}
\ead{andres.mendez@cinvestav.com}
\author[label2]{\uppercase{E.\ Rodriguez-Tello}}
\ead{ertello@cinvestav.com}

\affiliation[label1]{organization={Cinvestav Unidad Guadalajara},
            addressline={Av. del Bosque 1145}, 
            city={Zapopan},
            postcode={45017}, 
            state={Jalisco},
            country={Mexico}}

\affiliation[label2]{organization={Cinvestav Unidad Tamaulipas},
            addressline={Km. 5.5 Carretera Victoria-Soto La Marina}, 
            city={Victoria},
            postcode={87130}, 
            state={Tamaulipas},
            country={Mexico}}

\begin{abstract}
In precision agriculture, one of the most important tasks when exploring crop production is identifying individual plant components. There are several attempts to accomplish this task by the use of traditional 2D imaging, 3D reconstructions, and Convolutional Neural Networks (CNN).  However, they have several drawbacks when processing 3D data and identifying individual plant components. Therefore, in this work, we propose  a novel Deep Learning architecture to detect components of individual plants on Light Detection and Ranging (LiDAR) 3D Point Cloud (PC) data sets. This architecture is based on the concept of Graph Neural Networks (GNN), and feature enhancing with Principal Component Analysis (PCA). For this, each point is taken as a vertex and by the use of a K-Nearest Neighbors (KNN) layer, the edges are established, thus representing the 3D PC data set. Subsequently, EdgeConv layers are used to further increase the features of each point. Finally, Graph Attention Networks (GAT) are applied to classify visible phenotypic components of the plant, such as the leaf, stem, and soil.  This study demonstrates that our graph-based deep learning approach enhances segmentation accuracy for identifying individual plant components, achieving percentages above 80\% in the IoU average, thus outperforming other existing models based on point clouds.
\end{abstract}


\end{frontmatter}



\section{Introduction}
\label{intro1}

Accurate identification of plant components, such as leaves, stems, and soil regions, are a crucial task  in precision agriculture \cite{carroll_crop_2017}. Therefore, precise plant component identification allows for better biomass estimation \cite{zhang2020biomass},  targeted resource allocation \cite{fertilizantesagua},  crop productivity \cite{productividad1}, plant diseases discovery \cite{deteccenf}, and finally yield productivity estimation \cite{productividad2}.

In order to address the problem of plant component identification, Deep Learning Neural Networks are proposed to identify the primary structures of plants.
Deep Learning (DL) models have seen significant growth and evolution \cite{industry1, applicationagriculture1}, making them essential for point segmentation in DL novel projects.

A key advancement in Deep Learning has been the development of Convolutional Neural Networks (CNNs). Although CNNs have predominantly been applied to 2D image processing tasks \cite{cnn1}, extending them to handle 3D data is challenging due to the irregular and unordered nature of point clouds datasets. Unlike pixels in 2D images, which possess a uniform neighborhood structure, points in 3D PC have varying neighbor densities and distributions, which complicate the direct application of CNN models. Previous studies, such as the work by Bernhard et al \cite{japes_multi-view_2018, surveydeeplearning3D}, have explored methods to adapt conventional 2D CNN-based algorithms for effective processing of 3D data. In contrast, other novel models took a different approach by processing the point clouds directly. For example, the novel model PointNet \cite{qi2016pointnet} is a pioneer in this field. Unlike traditional convolutional models, it operates directly on raw point datasets without voxelization or projection. On the other hand, it employs symmetric aggregation functions such as max pooling and T-Net transformation to maintain permutation invariance. Therefore, PointNet learns global features while preserving the geometric structure in point cloud data.

However, PointNet has limitations in capturing fine-grained local structures, and the relationships among neighboring points, which are crucial for segmenting complex geometries. In order to overcome these shortcomings, other models have been proposed, such as Graph Convolutional Networks (GCN) \cite{gcn}. These models take the PC dataset and convert it into a graph, where each point acts as a vertex, and edges are defined by local neighborhood relationships. This representation allows for explicit modeling of both local and global dependencies through message passing, offering a more flexible framework for tasks that involve complex spatial structures \cite{PCsurvey}. In order to further improve the relationship between the points and their neighbors, another model such as Dynamic Graph CNN (DGCNN) \cite{DGCNN} enhances this idea by dynamically constructing graphs based on feature space, allowing the network to update neighborhood connections at each layer. This dynamic construction, combined with edge-based convolutions, has proven highly effective for point cloud segmentation tasks, offering a more flexible and spatially aware representation than fixed neighborhood approaches.

Building upon these advancements, Graph Attention Networks (GAT) \cite{GAT} introduce an attention mechanism into the graph-based learning paradigm, enabling the model to assign varying levels of importance to neighboring vertices during feature aggregation. Unlike GCN and DGCNN, which rely on uniform or implicitly learned weights for neighborhood aggregation, GAT explicitly compute attention coefficients for each edge in the graph. This allows the network to focus more on relevant points while down-weighting less informative ones. Such a mechanism is particularly beneficial in the context of point cloud segmentation, where certain regions may exhibit higher structural or semantic importance than others.
By incorporating attention-based weighting, GAT enable the model to capture complex spatial patterns more effectively, adapting dynamically to the  heterogeneity of plants point cloud data. A common limitation of these previous architectures is the insufficient explicit modeling of feature interactions among neighboring points, neglecting to fully exploit local geometric context.

\begin{figure*}[!t]
    \centering
    \includegraphics[width=\linewidth]{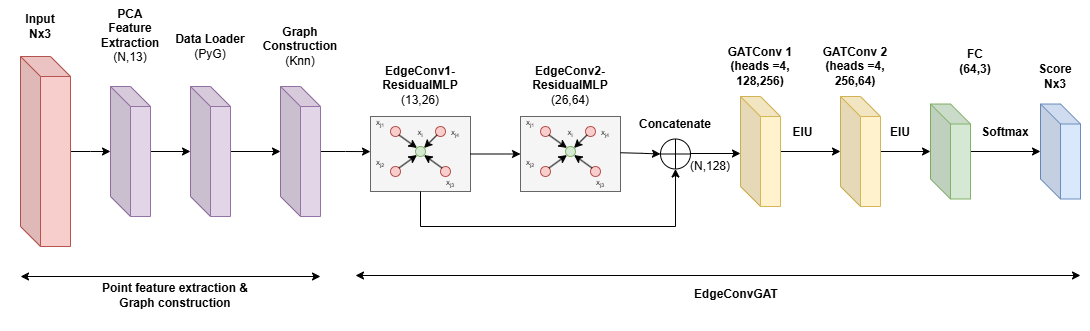}
    \caption{Our proposed model is a graph based EdgeGAT architecture. The first stage enhances point features by incorporating PCA-based features and constructing the graph representation. The second stage refines features using EdgeConv and leverages Graph Attention Networks for pretraining.}
    \label{fig1}
\end{figure*}

In order to address this challenge, we propose a novel architecture named EdgeGAT. The general architecture of this model can be seen in Figure \ref{fig1}, which leverages graph-based processing for enhanced point-cloud segmentation. For this, the proposed model applies Principal Component Analysis (PCA) as a preprocessing step. This step extracts geometric components present in the raw PC such as normals, curvature, planarity, and others; this also reduces noise and redundancy, but also improves the subsequent neighborhood construction and feature extraction phases. Following this, local geometric features are extracted and expanded using EdgeConv layers, which aggregate information from neighboring points while capturing directional edge relationships. These enriched features are then passed to a hierarchy module composed of GAT layers. These layers apply attention mechanisms over the constructed graph to selectively focus on the most relevant neighbors during feature propagation. This hierarchical refinement boosts the model ability to generalize across complex geometries.

The remainder of this paper is divided into the following sections: Section II discusses related work in point cloud segmentation and graph-based methods. Section III presents an overview of graph-based architectures used in this study. Section IV describes our proposed EdgeGAT model in detail. Section V outlines the experimental setup and presents the obtained results. Finally, Section VI offers conclusions and directions for future work.

\section{Previous Work}

In this section, we review the theoretical foundations and recent advances related to graph-based methods for PC segmentation. We first introduce fundamental graph concepts and their applicability to irregularly structured data. Subsequently, we discuss several prominent deep-learning architectures specifically designed for graph-structured data, including both classic and state-of-the-art approaches. 

\subsection{Graphs}
Graphs are versatile and powerful structures that enable the modeling of complex relationships between data in various applications \cite{graphs1}. Each graph consists of vertices and edges that funge as connections between them. In the context of point cloud processing, graphs offer an efficient way to preserve the topology details of the dataset. Translating a point cloud into a graph enables capturing local relationships between points. For this, graphs are a good option to represent irregular grids data. Thus, rather than 2D pixels, 3D PC can present a complex structure, specially when leaf and stem on a plant-based PC show.  With this in mind, we can construct a graph from our PC, given that a graph is represented as $G = (V,E)$
 where $|V|$ denotes the set of vertices and $E$ the set of edges. Therefore, we take as a vertex each point of the sub-sampled PC. With each point on the PC acting as a vertex, the final step is to create an edge for each point using the Euclidean distance between points. This can be performed using the K-nearest neighbor method, which connects two vertices $p, q$ by an edge if the distance is the k-th smallest distance from $p$ to any other point from PC. These edges can be represented with an adjacency matrix, which indicates the presence or absence of an edge between a pair of vertices. 

\subsection{Graph-Based Architectures}

In order to address the segmentation task on 3D point clouds, we consider six different deep learning architectures based on graph or point representations: (1) Graph Attention Network \cite{GAT}, (2) Graph Convolutional Network (GCN) \cite{gcnkipf}, (3) PointNet \cite{qi2016pointnet}, (4) Dynamic Graph CNN (DGCNN) \cite{DGCNN}, (5) GCN-UNet \cite{gcnunet2019}, and (6) our proposed model, EdgeGAT. These models are selected for their strong performance and conceptual relevance in processing unordered point clouds or graphs. In the following subsections, we briefly describe each model, highlighting their key components and main mathematical formulation.

\subsubsection{Graph Attention Networks}
Graph Attention Networks introduce the attention mechanism to graph-structured data, allowing each vertex to weigh the importance of its neighbors during the message-passing process. This enables the model to dynamically adjust the influence of different neighbors, which is particularly useful in irregular and sparse structures like point clouds. The main operation in GAT can be described as:

\[
\mathbf{h}_i' = \sigma\left(\sum_{j \in \mathcal{N}(i)} \alpha_{ij} \mathbf{W} \mathbf{h}_j\right),
\]

\noindent where:
 $\mathbf{h}_i$ is the input feature vector of vertex \(i\), and $\mathbf{h}_j$ is vector of neighbor \(j\),
 \(\mathbf{W}\) is a shared weight matrix, $\mathbf{h'_i}$ is the new node layer.
 \(\alpha_{ij}\) is the attention coefficient between vertices \(i\) and \(j\), calculated as:
\[
\alpha_{ij} = \frac{\exp\left(\text{LeakyReLU}\left(\mathbf{a}^\top [\mathbf{Wh}_i \Vert \mathbf{Wh}_j]\right)\right)}{\sum_{k \in \mathcal{N}(i)} \exp\left(\text{LeakyReLU}\left(\mathbf{a}^\top [\mathbf{Wh}_i \Vert \mathbf{Wh}_k]\right)\right)}
\]

\subsubsection{Graph Convolutional Networks (GCN)}
GCN generalizes the concept of convolution to non-Euclidean data by aggregating information from neighboring vertices. In its basic form, each vertex updates its representation by averaging (or summing) the features of its neighbors, followed by two operations:  a linear transformation and an activation function operation. The propagation rule for a GCN layer is:

\[
\mathbf{H}^{(l+1)} = \sigma\left(\tilde{\mathbf{D}}^{-\frac{1}{2}} \tilde{\mathbf{A}} \tilde{\mathbf{D}}^{-\frac{1}{2}} \mathbf{H}^{(l)} \mathbf{W}^{(l)}\right),
\]

\noindent Where:
 \(\tilde{\mathbf{A}} = \mathbf{A} + \mathbf{I}\) is the adjacency matrix with added self-loops,
 \(\tilde{\mathbf{D}}\) is the degree matrix of \(\tilde{\mathbf{A}}\),
 \(\mathbf{W}^{(l)}\) is the trainable weight matrix for layer \(l\), and  
 \(\sigma\) is a nonlinear activation function (e.g., ReLU).

\subsubsection{PointNet}
PointNet was one of the first architectures designed to directly process raw 3D point clouds without converting them into voxel grids or meshes. It treats each point independently using shared MLPs, followed by a global symmetric aggregation function (max pooling) to ensure permutation invariance.

\[
f(\{x_1, ..., x_n\}) \approx g\left(\underset{i=1,\ldots,n}{\text{MAX}}\left(h(x_i)\right)\right),
\]

\noindent where: 
\(n\) is the number of points in the PC,
 \(h\) is a point-wise feature extractor (MLP),
 \(\text{MAX}\) is a symmetric function that aggregates features across all points, and
 \(g\) is a final MLP for classification or segmentation.

\subsubsection{Dynamic Graph CNN (DGCNN)}
DGCNN extends GCNs by dynamically updating the graph structure at each layer based on feature space distances rather than fixed spatial proximity. This allows the model to adapt its neighborhood structure as feature representations evolve. Its core operation is the \textbf{EdgeConv}, defined as:

\[
\mathbf{h}_{i}^{(l+1)} = \underset{j \in \mathcal{N}(i)}{\text{MAX}} \ \phi\left(\mathbf{h}_i^{(l)}, \mathbf{h}_j^{(l)} - \mathbf{h}_i^{(l)}\right),
\]

\noindent where:
 \(\phi\) is a learnable function (usually an MLP),
 \(\mathcal{N}(i)\) denotes the k-nearest neighbors of point \(i\),
and edge features are dynamically recomputed at each layer.

\subsubsection{GCN-UNet}
GCN-UNet follows a U-shaped architecture inspired by UNet in CNNs, integrating pooling and unpooling operations adapted to graph data. It combines local context preservation with hierarchical abstraction, making it suitable for segmentation tasks where spatial consistency is important.

Its downsampling uses graph pooling techniques such as top $k$ pooling, while upsampling is done through interpolation or learned upsampling methods over the graph structure. The main idea behind GCN-UNet is to hierarchically encode and decode graph features:

\begin{align*}
\text{Encoder: } &\quad 
  \mathbf{H}^{(l)}
  \;\xrightarrow{\;\text{Pool}\;}
   \text{Pool} \hspace{0.1cm}\!\bigl(\mathbf{H}^{(l)}\bigr)
  \;\xrightarrow{\;\text{GCN}\;}
  \mathbf{H}^{(l+1)},
  \\[4pt]
\text{Decoder: } &\quad 
  \mathbf{H}^{(l)}
  \;\xrightarrow{\;\text{Unpool}\;}
   \text{Unpool} \hspace{0.1cm} \!\bigl(\mathbf{H}^{(l)}\bigr)
  \;\xrightarrow{\;\text{GCN}\;}
  \mathbf{H}^{(l+1)}.
\end{align*}

\subsubsection{EdgeGAT}

Our proposed model, EdgeGAT, is designed to leverage both the edge-level geometric relationships from EdgeConv and the adaptive feature weighting of attention mechanisms from GAT. Inspired by DGCNN \cite{DGCNN}, we first construct a dynamic graph based on $k$ nearest neighbors in the feature space. Then, we apply an attention-based convolution that operates over vertex features. The EdgeConv operation captures local geometric context through the function:

\[
\mathbf{e}_{ij} = h_{\theta}(\mathbf{h}_i, \mathbf{h}_j - \mathbf{h}_i),
\]

\noindent where:
- \(\mathbf{h}_i\) and \(\mathbf{h}_j\) are the features of vertices \(i\) and \(j\),
- \(h_{\theta}\) is a learnable MLP. We integrate attention weights \(\alpha_{ij}\) into this formulation as:

\[
\mathbf{h}_i' = \sum_{j \in \mathcal{N}(i)} \alpha_{ij} \cdot \mathbf{e}_{ij} ,
\]

This allows the model to both learn geometric structure (via edge features) and adaptively weight neighbors (via attention), enhancing the expressive power for point cloud segmentation tasks.

\section{Proposed graph-based EdgeGAT model}

We propose a novel architecture, called EdgeGAT, that integrates enhanced point features with graph-based attention mechanisms to classify points in PC datasets based on previously labeled point cloud data.  This architecture is designed around the following concepts:
\begin{enumerate}
    \item A point feature extraction and generation, 
    \item A graph extraction for KNN,
    \item EdgeConv layers for vertex representation,
    \item Graph Attention layers for global and local context,
    \item An explanation on the use of a residual MLP.
\end{enumerate}

Finally, we have a section to describe the training of this entire architecture.

\subsection{Point Feature Extraction}
\label{sec:PCA}

Raw point cloud data, initially represented by only the 3D spatial coordinates $(x, y, z)$, lacks sufficient geometric context for complex semantic tasks. To enrich this representation, we incorporate a series of local geometric descriptors derived from Principal Component Analysis (PCA) computed on each point’s local neighborhood \cite{hotelling1933pca, pearson1901pca, jolliffe2002pca, weinmann2015localPCA}.

The final input vector for each point expands from 3 to 13 dimensions, incorporating the following features:

\begin{itemize}
    \item Coordinates: The original input with 3 channels $(x, y, z)$
    \item Normals: \[
\mathbf{C} = \frac{1}{k} \sum_{i=1}^{k} (\mathbf{p}_i - \bar{\mathbf{p}})(\mathbf{p}_i - \bar{\mathbf{p}})^T,
\]

\noindent where $\mathbf{p}_i$ denotes the $i$-th neighboring point and $\bar{\mathbf{p}}$ is the centroid of the neighborhood. The eigenvector corresponding to the smallest eigenvalue $\lambda_0$ of $\mathbf{C}$ defines the estimated surface normal at point $\mathbf{p}$ \cite{covariancenormals1}.

\item Curvature: A measure of local surface variation, defined as:

\[
\text{Curvature} = \frac{\lambda_0}{\lambda_0 + \lambda_1 + \lambda_2},
\]

where $\lambda_0 \leq \lambda_1 \leq \lambda_2$ are the eigenvalues of the covariance matrix. 

\item Linearity: Indicates how well points align along a line in the neighborhood. Defined as:

\[
\text{Linearity} = \frac{\lambda_2 - \lambda_1}{\lambda_2},
\]

a value close to 1 indicates strong alignment along one direction (e.g., stems), while lower values suggest more isotropic or planar regions.

\item Planarity: Measures how well the local neighborhood fits a plane. Defined as:

\[
\text{Planarity} = \frac{\lambda_1 - \lambda_0}{\lambda_2},
\]

high planarity values suggest that the points are spread over a 2D surface, such as plants components.

\item Scaterring: Reflects how dispersed the local neighborhood is in 3D space. It is computed as:

\[
\text{Scattering} = \frac{\lambda_0}{\lambda_2},
\]

a low value indicates structured data (e.g., lines or planes), while a high value suggests volumetric or noisy regions.

\item Omnivariance: A measure of volumetric dispersion in 3D space, calculated as the geometric mean of the eigenvalues:

\[
\text{Omnivariance} = (\lambda_0 \cdot \lambda_1 \cdot \lambda_2)^{1/3},
\]

higher values represent more volumetric or isotropic neighborhoods.

\item Anisotropy: Measures the degree of directionality in the local neighborhood. Computed as:

\[
\text{Anisotropy} = \frac{\lambda_2 - \lambda_0}{\lambda_2},
\]

higher values indicate strong alignment along a principal direction, such as stems or elongated structures.

\item Eigenentropy: Describes the entropy of the local eigenvalue distribution, capturing the complexity or randomness of point dispersion. Defined as:

\[
\text{Eigenentropy} = -\sum_{i=0}^{2} \tilde{\lambda}_i \log(\tilde{\lambda}_i)
\quad \text{where} \quad \tilde{\lambda}_i = \frac{\lambda_i}{\lambda_0 + \lambda_1 + \lambda_2},
\]

where higher values suggest more uniform distribution of variance, while lower values imply strong directional structure.
\end{itemize}

Here, $\lambda_0 \leq \lambda_1 \leq \lambda_2$ are the eigenvalues obtained by performing PCA over the covariance matrix of the point local neighborhood. These values describe the spatial distribution and variation of neighboring points, offering geometric intuition such as flatness (planarity), elongation (linearity), or compactness (scattering) \cite{planilinea1}.

This 13-dimensional feature vector enables the model to distinguish subtle structural differences between different plant components, such as stems, leaves, and surrounding soil. The inclusion of curvature and entropy, for instance, helps in capturing surface smoothness and complexity, which are useful for clasification in semantic segmentation tasks like in plants datasets \cite{weinmann2015localPCA, hotelling1933pca} .

\subsection{Edge and graph construction}

Following the feature extraction step, each point in PC is now represented by a 13-dimentional feature vector, encompassing not only the original spatial coordinates \((x, y, z)\) but also a set of geometric additional features derived from PCA feature generation. Therefore, in order to capture local geometric relationships between points, we construct a graph using the KNN algorithm in the spatial features. 

The graph construction process considers only the 3D spatial position of each point, denoted as \( \mathbf{x}_i \in \mathbb{R}^3 \), for neighbor identification. In order to construct the edges of the graph, we use the Euclidean distance between two points defined as:
\begin{equation}
    d(\mathbf{p}_i, \mathbf{p}_j) = \left\lVert \mathbf{p}_i - \mathbf{p}_j \right\rVert_2,
\end{equation}
where \(\left\lVert \cdot \right\rVert_2\) is the \(\ell_2\)-norm. 
For each point \(\mathbf{p}_i\), its \(k\)-nearest neighbors are determined by
\begin{equation}
    \mathcal{N}_k(\mathbf{p}_i) = \left\{ \mathbf{p}_j \;\middle|\; j \in \arg\min_{j \neq i} d(\mathbf{p}_i, \mathbf{p}_j), \; |\mathcal{N}_k| = k \right\}.
\end{equation}

Once the neighborhood is identified, we construct an undirected edge between \( \mathbf{p}_i \) and each of its neighbors \( \mathbf{p}_j \in \mathcal{N}_k(\mathbf{x}_i) \). The graph is encoded via an edge index matrix \( \mathbf{E} \in \mathbb{R}^{2 \times |E|} \), where each column defines a pair of connected vertex indices. While the edge structure is based on spatial proximity, the 13-dimensional features associated with each vertex are described as  \( \{\mathbf{p}_1, \mathbf{p}_2, \dots, \mathbf{p}_N\} \subset \mathbb{R}^{13} \), and serve as the vertex attributes used in downstream processing stages such as EdgeConv and Graph Attention layers. This method is well-established for pattern recognition and graph construction \cite{cover1967nearest}
 and is performed by other applications such as Pointnet++ \cite{pointnet2} which leverage KNN to model local geometric relationships in point clouds. 

\subsection{Residual MLP for Enhanced Feature Propagation}
\label{sec:residual_mlp}

In our architecture, rather than using a conventional Multi-Layer Perceptron in the EdgeConv blocks, we employ a \textbf{Residual MLP}. A standard MLP performs a sequence of linear transformations followed by non-linear activations. For instance, a two-layer MLP can be expressed as:

\begin{equation}
    \mathbf{y} = \sigma\bigl(\mathbf{W}_2 \, \sigma(\mathbf{W}_1 \mathbf{x} + \mathbf{b}_1) + \mathbf{b}_2\bigr),
\end{equation}

\noindent where \(\mathbf{x}\) is the input vector, \(\mathbf{W}_1, \mathbf{W}_2\) are learnable weight matrices, \(\mathbf{b}_1, \mathbf{b}_2\) are biases, and \(\sigma(\cdot)\) denotes a non-linear activation function, specifically Leaky ReLU in our implementation.

While standard MLPs perform adequately in shallow configurations, they can suffer from vanishing gradients and degraded performance in deeper networks due to the loss of original input information through successive transformations. To address these limitations, we adopt the \textbf{Residual MLP}, inspired by the residual learning paradigm introduced by He et al.~\cite{he2016deep}. This approach employs a skip (residual) connection that directly propagates the input signal to the output, facilitating improved gradient flow and better feature preservation. Residual learning has demonstrated superior performance in various deep graph neural network architectures~\cite{rong2020dropedge, RMLP2, RMLP3}.
 Formally, given an input feature vector \(\mathbf{x} \in \mathbb{R}^{d_{\text{in}}}\), the Residual MLP computes:
\begin{equation}
    \mathbf{y} = \mathcal{F}(\mathbf{x}) + \mathcal{S}(\mathbf{x}),
\label{eq:skip_connect}
\end{equation}

\noindent where \(\mathcal{F}(\mathbf{x})\) denotes the sequence of linear transformations, batch normalization, Leaky ReLU activations, and dropout layers. The skip connection, \(\mathcal{S}(\mathbf{x})\), is either a linear projection \(\mathbf{W}_s \in \mathbb{R}^{d_{\text{out}} \times d_{\text{in}}}\) if input and output dimensions differ (\(d_{\text{in}} \neq d_{\text{out}}\)), otherwise, in the case of the same input and output dimensions (\(d_{\text{in}} =d_{\text{out}}\)), it returns the initial vector.

In the context of EdgeConv implementation, described subsequently in Section~\ref{sec:edgeconv}, the Residual MLP consists specifically of two hidden layers, expanding input feature dimensions sequentially. This facilitates progressive feature refinement while ensuring stable training through residual connections.

\subsection{EdgeConv Integration for Graph vertex Features}
\label{sec:edgeconv}

\begin{figure}
    \centering
    \includegraphics[width=1.0 \linewidth]{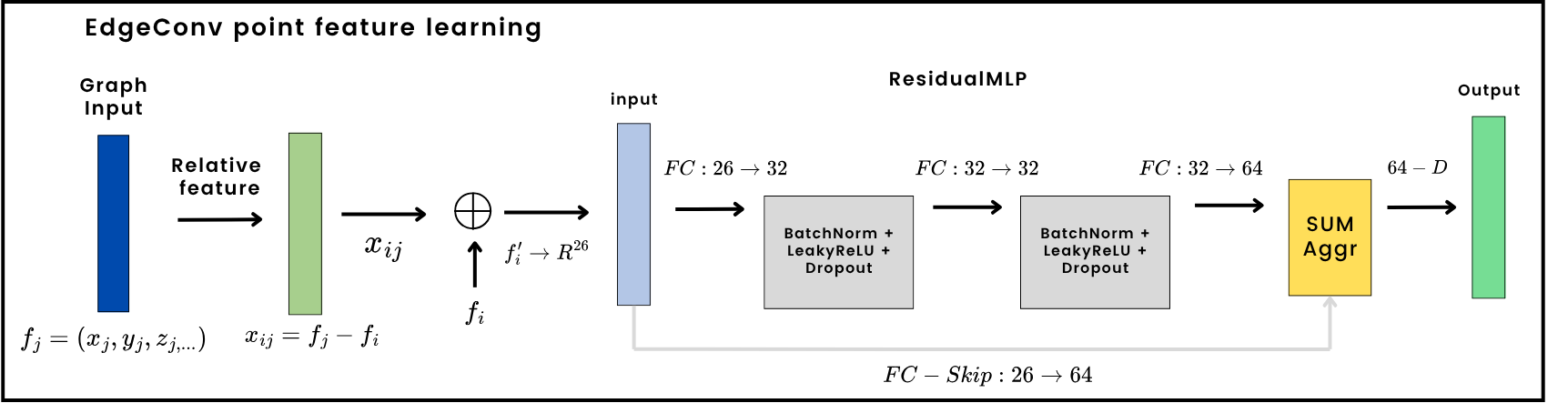}
    \caption{Architecture of the EdgeConv module for feature enhancement. The Residual MLP plays a key role in refining features after concatenation with the original input, resulting in an enhanced feature representation. }
    \label{fig:edgeConvSA}
\end{figure}

Building upon the Residual MLP previously described, our proposed architecture integrates local geometric information into vertex representations using the \textbf{EdgeConv} operator, initially introduced by Wang et al.~\cite{wang2019dynamic} for point cloud analysis. EdgeConv enriches vertex embeddings by explicitly capturing both absolute and relative geometric features from each vertex and its neighbors, as illustrated in Figure~\ref{fig:edgeConvSA}.

Formally, given a vertex \( v_i \) with an initial enriched feature vector \(\mathbf{f}_i \in \mathbb{R}^{13}\), EdgeConv computes edge features for each neighbor edge \( (i, j) \in E \) as follows:

\begin{equation}
    \mathbf{e}_{ij} = \Phi\Bigl([\mathbf{f}_i \,\, \| \,\, \mathbf{f}_j - \mathbf{f}_i]\Bigr),
\end{equation}

\noindent where \( \Phi(\cdot) \) is the Residual MLP described previously, and \( \| \) denotes concatenation.

Thus, we apply two consecutive EdgeConv blocks to progressively refine vertex embeddings:

\begin{itemize}
    \item \textbf{First EdgeConv block:} Initially, the relative neighbor differences \(\mathbf{f}_j - \mathbf{f}_i \in \mathbb{R}^{13}\) are computed, capturing local geometric context around each vertex. These differences are concatenated within the vertex original features \(\mathbf{f}_i\), producing an enhanced 26-dimensional edge vector (\(\mathbf{x}_{ij} = [\mathbf{f}_i \| \mathbf{f}_j - \mathbf{f}_i]\in \mathbb{R}^{26}\)). This vector is subsequently processed by the ResidualMLP, expanding dimensions from 26 to 32, and then from 32 to 64 dimensions through a simple MLP. After applying sum aggregation  over neighboring edges, the first refined vertex embedding \(\mathbf{h}_i^{(1)} \in \mathbb{R}^{64}\) is obtained.

    \item \textbf{Second EdgeConv block:} The refined vertex embeddings \(\mathbf{h}_i^{(1)} \in \mathbb{R}^{64}\) from the previous block serve as input. Following the same methodology, relative differences are computed as \(\mathbf{h}_j^{(1)} - \mathbf{h}_i^{(1)}\), producing vectors of 64 dimensions. Concatenation of these relative differences with the vertices own features (\(\mathbf{h}_i^{(1)}\)) yields 128-dimensional edge vectors, which are subsequently processed by a second Residual MLP, this time reducing dimensionality from 128 to 32 and then expanding back to 64. After mean aggregation over neighbors, we obtain the second refined embeddings, \(\mathbf{h}_i^{(2)} \in \mathbb{R}^{64}\).
\end{itemize}

Finally, to preserve the initial geometric information and facilitate feature reuse, we concatenate the original input vector \(\mathbf{f}_i\) with the output of the second EdgeConv  \(\mathbf{h}_i^{(2)}\), yielding a final vertex representation:

\begin{equation}
    \mathbf{h}_i^{\text{final}} = [\mathbf{f}_i \,\, \| \,\, \mathbf{h}_i^{(2)}] \in \mathbb{R}^{77}.
    \label{eq:edgeconv_final}
\end{equation}

\noindent This enriched embedding captures both low-level geometric structure and high-level edge-based contextual information. These representations then serve as input for subsequent graph-attention layers, as detailed in the following section.

\subsection{ Graph Attention layers }
Following the EdgeConv stages, each point in the graph is represented by a feature vector of dimension \(77\). This dimensionality arises from the concatenation of the original 13 input features with the outputs of the two successive EdgeConv blocks, 26 and 64 dimensions respectively, with partial overlaps due to shortcut connections. These enriched representations serve as input to our GAT, which is designed to further refine vertex features by capturing both local and global dependencies through attention mechanisms \cite{GAT, Attention1}.

In each GAT layer, a shared linear transformation \(\mathbf{W}\) is applied to every vertex feature vector \(\mathbf{h}_i \in \mathbb{R}^{77}\), yielding a transformed feature space. To model the importance of neighboring vertices, the attention coefficient between vertex \(i\) and its neighbor \(j\) is computed as:
\begin{equation}
    \alpha_{ij} = \text{softmax}_j\left( \sigma\left( \mathbf{a}^\top \left[ \mathbf{W}\mathbf{h}_i \parallel \mathbf{W}\mathbf{h}_j \right] \right) \right),
\end{equation}
where \(\mathbf{a}\) is a learnable weight vector and \(\sigma\) is a Leaky ReLU activation function. These coefficients quantify the attention  that vertex \(i\) assigns to each of its neighbors.

The vertex feature update is then performed through an attention-weighted aggregation of neighbor features:
\begin{equation}
    \mathbf{h}'_i = \sum_{j \in \mathcal{N}(i)} \alpha_{ij} \, \mathbf{W}\mathbf{h}_j.
\end{equation}

This attention mechanism enables the model to selectively focus on informative neighbors while down-weighting less relevant ones. In our implementation, two stacked GAT layers are used, allowing the network to progressively capture richer features at different levels of neighborhood abstraction.

\subsection{Training with Graph Attention Networks}
After enriching the point features to a 77-dimensional representation with Residual MLPs and EdgeConv, we employ GAT to perform vertex classification on the constructed graph \( G = (V, E) \). Each vertex in the graph represents a point in the point cloud and is associated with a 77-dimensional feature vector. 

Unlike traditional GCNs, which use uniform aggregation of neighboring features, GAT layers incorporate a learnable attention mechanism to assign different weights to neighboring vertices. This allows the model to emphasize relevant context while maintaining spatial adaptability. As introduced by Veličković et al. \cite{GAT}, each attention head independently processes the neighborhood of a vertex, capturing diverse relational patterns.

The first GAT layer receives an input of shape \( N \times 77 \), where \( N \) is the number of vertices, and produces a higher-dimensional representation using four parallel attention heads, each with 64 hidden units. The outputs of these heads are concatenated, resulting in a representation of shape \( N \times 256 \):

\[
\mathbf{H}_1 = \text{GAT}_{\text{concat}}\left( \mathbf{X}, \mathbf{E} \right) \in \mathbb{R}^{N \times 256}
\]

\noindent where \( \mathbf{X} \in \mathbb{R}^{N \times 77} \) is the input feature matrix and \( \mathbf{E} \) the set of edges, constructed via a fixed \( k \)-nearest neighbor strategy with \( k = 16 \).

A second GAT layer then processes this intermediate representation. Again using four attention heads, but this time averaging their outputs, the resulting representation has shape \( N \times 64 \):

\[
\mathbf{H}_2 = \text{GAT}_{\text{avg}}\left( \mathbf{H}_1, \mathbf{E} \right) \in \mathbb{R}^{N \times 64}
\]

\noindent A final fully connected layer projects this output into the classification space of three classes:

\[
\mathbf{Y} = \text{Linear}(\mathbf{H}_2) \in \mathbb{R}^{N \times 3}
\]

During training, the model minimizes a cross-entropy loss over all vertices. Optimization is performed using the Adam algorithm \cite{adam1} with a learning rate of 0.001, for 100 epochs, and a batch size of 8. A fixed graph structure with \( k = 16 \) neighbors is used throughout, though the architecture allows for extension to dynamic graph constructions based on spatial radius.

This attention-based architecture effectively leverages the rich 77-dimensional vertex features and the expressiveness of multi-head attention to capture both fine-grained and global contexts in complex plant structures.

\section{Experiments}
The objective of these experiments is to classify each vertex with our model into one of three predefined categories: soil, stem, or leaf. For this, each step carried out with our dataset is described below.

\subsection{Data}
The data was obtained from the work of the authors of \cite{cropmaize}, who allowed us to use the point clouds they had collected on corn. We refer to this dataset as Ao dataset in reference to its first author. These point clouds were acquired using terrestrial LiDAR. We also used a second dataset, this was obtained from Pheno4D \cite{Pheno4D} which had labeled point clouds of corn and tomato plants, of which we focused on the corn dataset, we kept the original Pheno4D name for analysis.

Each individual corn plant file originally contained over 50{,}000 points, a number too large to be directly processed. To address this, we applied a data augmentation step consisting of random jittering, rotation, translation, and scaling to enhance data variability and model robustness. After augmentation, uniform random downsampling was performed to reduce the number of points to 1{,}024 prior to training in both datasets. This sampling was done using the \texttt{Open3D} library, which selects a random subset of point indices without replacement.

Mathematically, this corresponds to selecting a subset $S \subset P$ such that $|S| = 1024$, where $P$ denotes the augmented set of points, and the selection is governed by a uniform probability distribution over the indices. For the PCA-based feature extraction process, we generated various geometric descriptors discussed in section \ref{sec:PCA}. Throughout this document, these descriptors are denoted by specific acronyms for convenience, as detailed in Table~\ref{tab:feature_acronyms}.

\begin{table}[h!]
\centering
\caption{Acronyms used for PCA-derived geometric feature sets}
\label{tab:feature_acronyms}
\begin{tabular}{ll}
\toprule
\textbf{Acronym} & \textbf{Feature Combination} \\ 
\midrule
XYZ & Raw 3D coordinates \\[2pt]
N & Surface normals \\[2pt]
C & Curvature \\[2pt]
L & Linearity \\[2pt]
P & Planarity \\[2pt]
S & Scattering \\[2pt]
O & Omnivariance \\[2pt]
A & Anisotropy \\[2pt]
E & Eigenentropy \\[2pt]
\bottomrule
\end{tabular}
\end{table}

\subsection{Data preprocessing}
The points used for classification are manually selected and visualized using MeshLab an Open-Source Mesh Processing Tool \cite{meshlab}. Once the dataset is fully prepared, the model can be trained to perform these classifications with our model. The distribution of the three classes is given by Table \ref{table1}.

\renewcommand{\arraystretch}{1.2}
\begin{table}[h!]
\centering
\caption{Table 1: Distribution of classes in Ao maize dataset (\%)}
\begin{tabular}{lccc}
\toprule
\textbf{} & \textbf{Leaves (\%)} & \textbf{Soil (\%)} & \textbf{Stem (\%)} \\ 
\midrule
Maize train model     & 50.15               & 29.95              & 19.91                \\
Maize test model      & 53.36               & 28.34              & 18.30                \\
Maize valid model     & 53.82               & 24.42              & 21.77                \\ 
\bottomrule
\end{tabular}
\label{table1}
\end{table}

\begin{figure}
    \centering
    \includegraphics[width=0.8\linewidth]{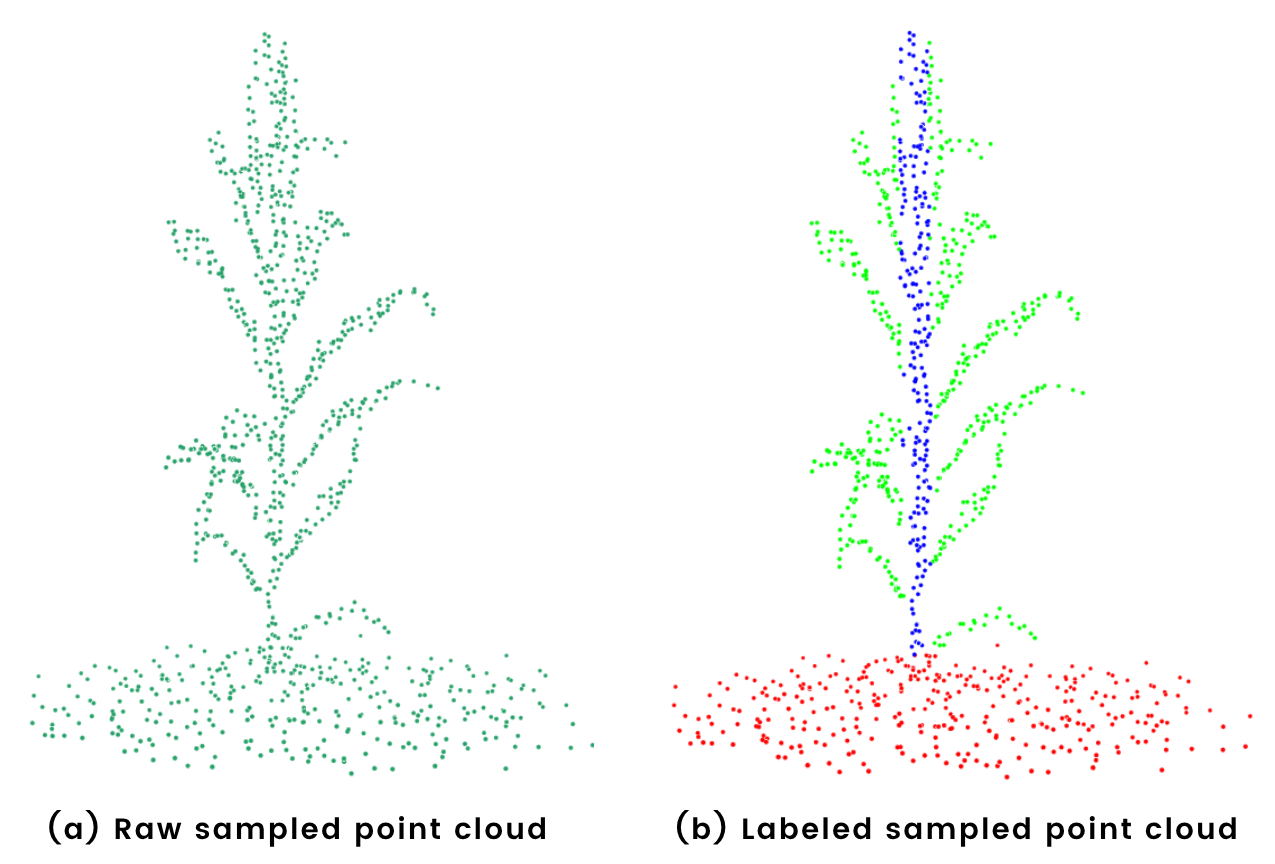}
    \caption{Three-class point cloud representation of a maize plant in Ao dataset.(a)  The raw point cloud data, (b) The classified point cloud, where the stem is labeled in blue, the leaves in green, and the soil in red.}
    \label{fig:Aolabel}
\end{figure}

\begin{figure}
    \centering
    \includegraphics[width=0.8\linewidth]{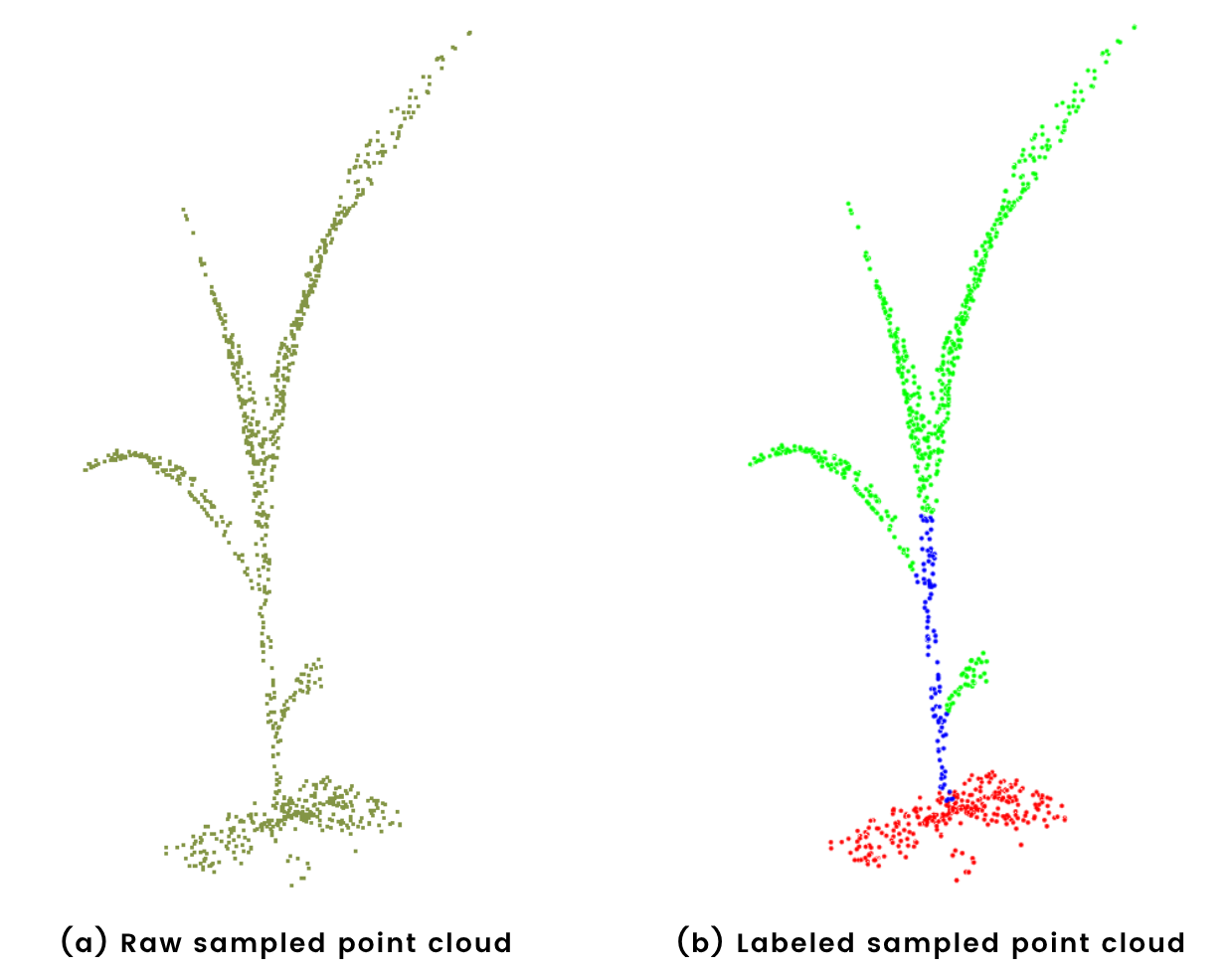}
    \caption{Three-class point cloud representation of a maize plant used for model training in Pheno4D dataset. (a) The raw point cloud data, (b) The classified point cloud, where the stem is labeled in blue, the leaves in green, and the soil in red.}
    \label{fig:Pheno4d label}
\end{figure}

Due to limited data access, the maize plants in the Ao dataset had an overall age of approximately 30 days and presented plant heights ranging from 80 cm to 130 cm. In contrast, the Pheno4D dataset contained younger maize plants aged between 4 and 10 days, with plant heights typically ranging from 10 cm to 30 cm. Both datasets exhibit significant structural variations, especially in leaf and stem configurations. Given the early developmental stage of plants in Pheno4D, the maize spike was grouped with the stem class, as it was not fully developed. Additionally, the upper and lower plant segments were combined into the stem category, simplifying classification. For clear visualization of the ground truth labels, distinct colors were assigned to represent each class: green for leaves, red for soil, and blue for stem. Examples of labeled maize plants from the Ao dataset are presented in Figure~\ref{fig:Aolabel}, while Figure~\ref{fig:Pheno4d label} illustrates labeling results for the Pheno4D dataset.

The classified datasets were evaluated using a five-fold cross-validation strategy. Specifically, each dataset was partitioned into five equally sized subsets. In each iteration, four folds, that contain 80\% of the data, are used to train the model, while the remaining fold is employed as a validation set for hyperparameter tuning and to detect potential overfitting. This procedure was repeated five times, ensuring each fold served exactly once as the validation set. The overall performance metrics were then calculated by averaging the results across all folds, thus providing a more robust model.

\subsection{Ablation study on feature enhancement with PCA-derived attributes}

Given the features obtained by PCA, we conducted a series of experiments, evaluating the impact of incorporating these features derived from PCA geometric descriptors. These experiments are performed principally in Ao dataset, and across four architectures: GCN, GAT, UNET, and UNET2. Where UNET and UNET2 are architectures based on the GCN following an encoder-decoder structure best described in Appendix A. Also, we employ a fixed graph construction strategy with \(k=16\).

\begin{figure}[h]
    \centering
    \includegraphics[width=1.00\linewidth]{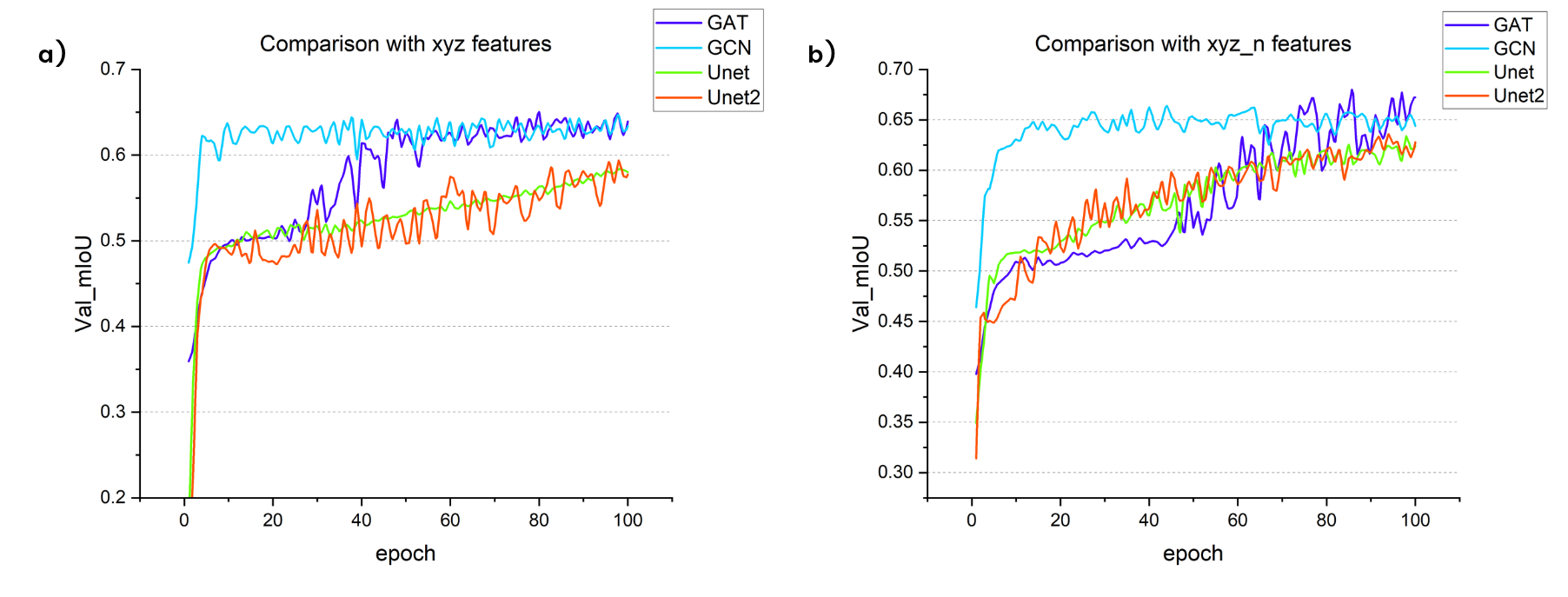}
    \caption{Validation mIoU results across different feature configurations for GCN, GAT, UNET, and UNET2 on Ao dataset. Each subplot shows the evolution of validation performance when including: (a) only xyz, (b) xyz + normals.}
    \label{fig:feature_expansion_xyn}
\end{figure}

For this, we start with the simplest feature configuration using only the spatial coordinates, and progressively include other descriptors such as: normals, curvature and normals and finally all our 10 PCA features. While the inclusion of curvature or normals individually led to only marginal improvements in validation mIoU. A more substantial boost is observed in Figure \ref{fig:feature_expansion_xyn} when combining both curvature and normal features. This is the reason behind the exploration of a richer 13-dimensional feature set derived from PCA.

\begin{figure}[h]
    \centering
    \includegraphics[width=1.00\linewidth]{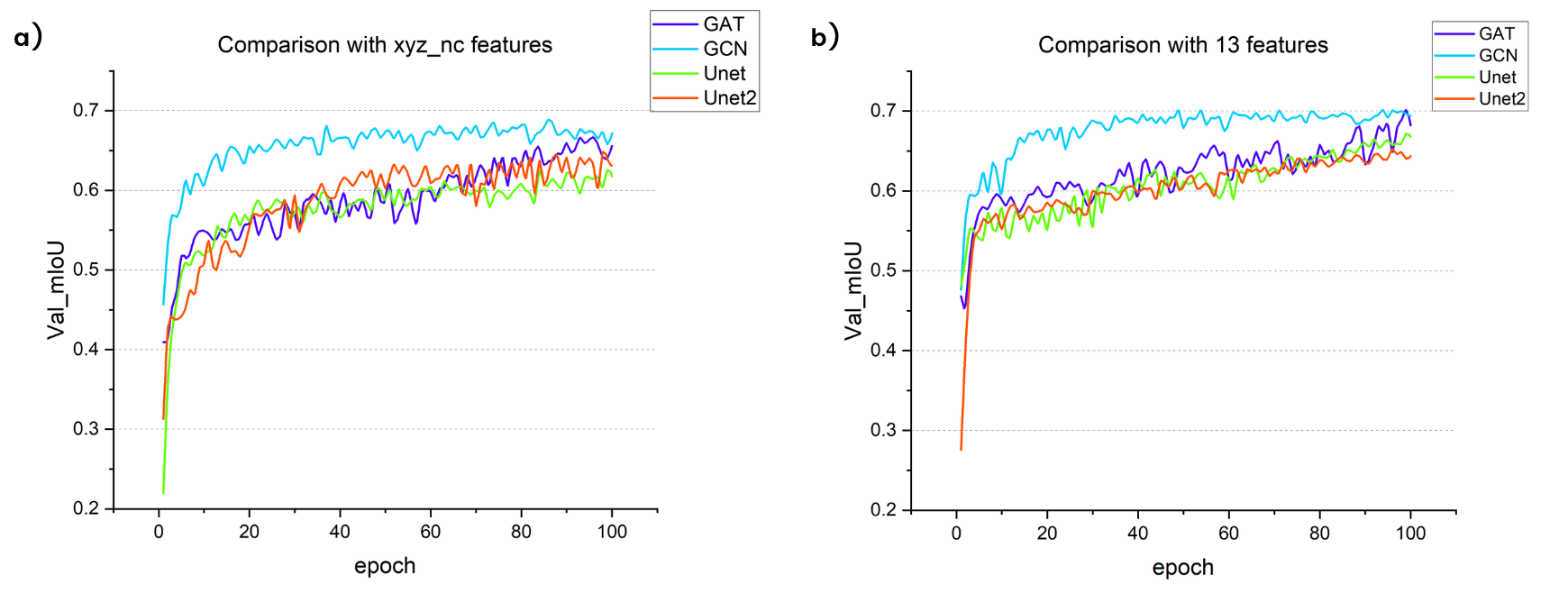}
    \caption{Validation mIoU results across different feature configurations for GCN, GAT, UNET, and UNET2 on Ao dataset. Each subplot shows the evolution of validation performance when including: (a) xyz + normals + curvature, (b) xyz + 10 PCA features.}
    \label{fig:feature_expansion_results}
\end{figure}

Figure~\ref{fig:feature_expansion_xyn}(a) presents the baseline performance using only the raw XYZ coordinates. Under this scenario, all four models exhibit comparable mIoU results after 60 epochs, with GAT and GCN showing similar performance, and both U-Net-based models converging below the 60\% threshold.

Figure \ref{fig:feature_expansion_xyn}(b) adding surface‐normal vectors to the XYZ coordinates, a notable performance differentiation emerges among the models. Here, the GAT model achieves the highest performance, followed closely by GCN. The two U-Net models display improvements as well, clustering their performances around the 65\% mark. Nevertheless, GAT consistently outperforms all other models in this configuration.

Expanding further on this analysis, Figure~\ref{fig:feature_expansion_results}(a) presents additional experiments. This figure includes curvature, alongside XYZ and normals, showing a clear positive trend across all models. The GCN model achieves the highest mIoU at approximately 67\%, followed closely by GAT, then U-Net2 and U-Net in that order. This demonstrates a beneficial impact from incorporating curvature as an additional feature, raising overall performance from roughly 60\% to 67\%.

Figure \ref{fig:feature_expansion_results} (b) evaluates the impact of using an expanded feature set comprising XYZ, normals, curvature, and the complete set of 10 PCA-derived features. Here, we observe further performance improvement, with GCN narrowly surpassing GAT, reaching a value slightly above 70\% after epoch 35. The GAT model achieves a similar level, but only after epoch 90. The two U-Net variants lag behind, remaining consistently below the 70\% mark. 


Given the previously observed improvements in model performance when integrating additional geometric features, we conduct an analysis comparing the training times of all evaluated architectures under the different feature configurations. Table \ref{tab:training_time_comparison} summarizes the time per epoch (in seconds) for each model and feature set. As seen, GCN and GAT architectures consistently exhibit significantly faster training speeds compared to the U-Net–based encoder–decoder models.

\renewcommand{\arraystretch}{1.2}
\begin{table}[h!]
\centering
\caption{Comparison of Training Times for Different Graph Architectures on Ao dataset}
\begin{tabular}{lcccc}
\toprule
\textbf{Features} & \textbf{GCN (s)} & \textbf{UNet (s)} & \textbf{UNet2 (s)} & \textbf{GAT (s)} \\
\midrule
XYZ & 12.04 & 31.24 & 28.97 & 11.23 \\
XYZ-N & 12.08 & 31.36 & 30.32 & 11.09 \\
XYZ-NC & 12.14 & 31.52 & 28.92 & 11.21 \\
XYZ-NCLPSOAE & 12.21 & 31.09 & 28.30 & 11.14 \\
\bottomrule
\end{tabular}
\label{tab:training_time_comparison}
\end{table}

The results highlight a clear efficiency advantage of GCN and GAT models. Specifically, GCN achieved consistently low training durations,  that have approximately 12.0 seconds per epoch across all feature configurations, while GAT maintained similar efficiency with epoch durations of around 11.0 seconds. Conversely, the U-Net based GCN models exhibited notably longer training times, ranging from approximately 28.3 to 31.5 seconds per epoch, depending on the specific feature set. This substantial efficiency gap, combined with the competitive or superior performance previously demonstrated, strongly supports the adoption of GNN-based architectures—particularly GCN and GAT—for this type of point cloud segmentation task.

In order to further evaluate the impact of adding 13 PCA-derived features with our model, we extended this comparison to include performance metrics such as validation loss and mIoU, as shown in Table~\ref{tab:training_time_comparison_allfeat}.

\renewcommand{\arraystretch}{1.2}
\begin{table}[h!]
\centering
\caption{Training Time and Performance Metrics Using 13 PCA Features in Ao dataset}
\begin{tabular}{lccc}
\toprule
\textbf{Model} & \textbf{Training Time (s)} & \textbf{mIoU (\%)} & \textbf{Val. Accuracy (\%)} \\ 
\midrule
GCN       &  12.25    &  67.33   &  79.20   \\
\textbf{GAT}       &  \textbf{11.70}    &  67.40   &  79.25   \\
UNet      &  25.07    &  64.07   &  77.53   \\
UNet2     &  30.92    &  65.92   &  79.50   \\
\textbf{EdgeGAT}   &  22.49    &  \textbf{73.44}&  \textbf{82.28}   \\
\bottomrule
\end{tabular}
\label{tab:training_time_comparison_allfeat}
\end{table}

\begin{figure}[h!]
    \centering
    \includegraphics[width=0.79\linewidth]{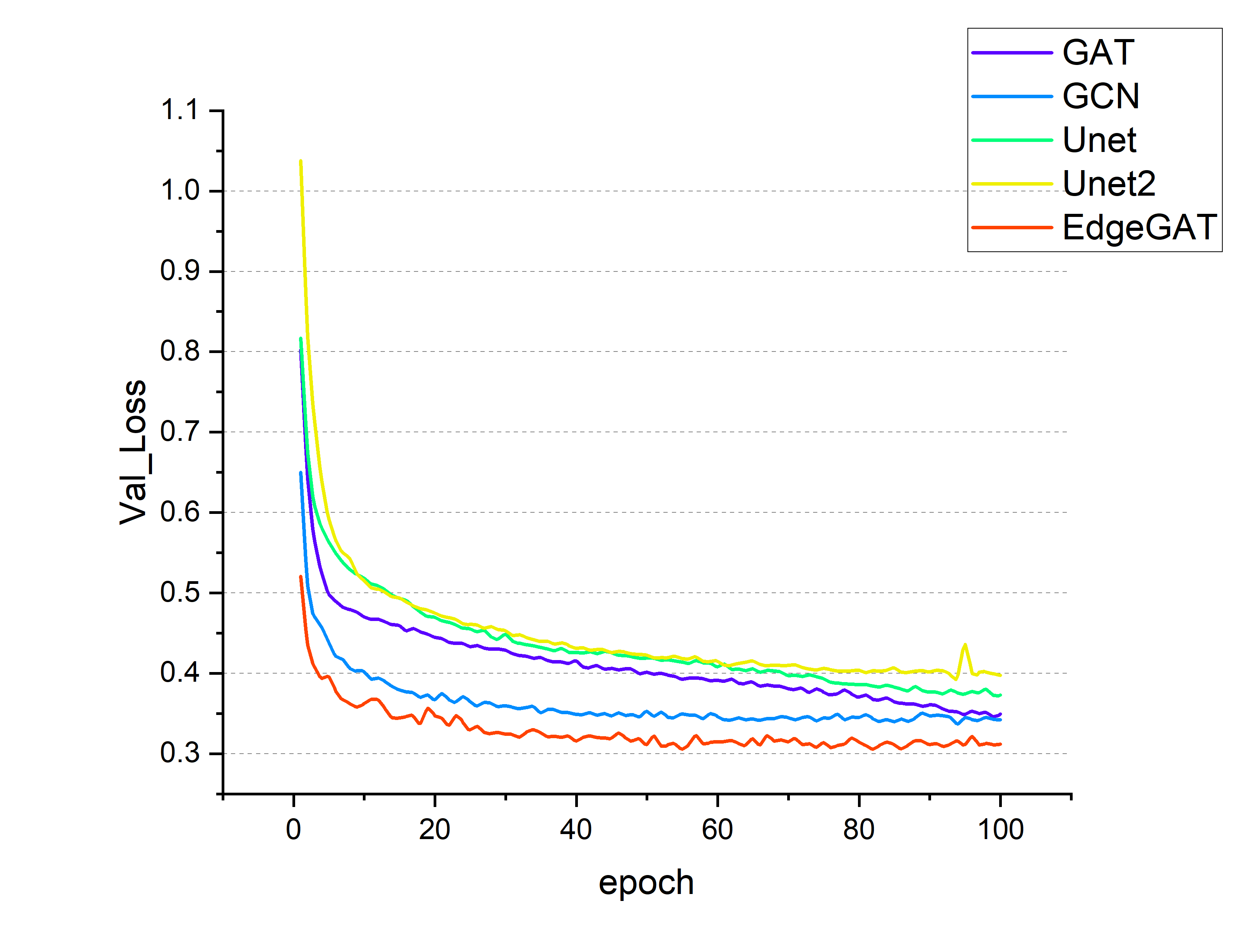}
    \caption{Validation Loss of GCN, GAT, UNet, UNet2, and EdgeGAT using 13 PCA-extracted features in Ao dataset.
}
    \label{fig:PCAimproveloss}
\end{figure}    

As shown in Table~\ref{tab:training_time_comparison_allfeat}, the GAT model exhibited relatively short training times while maintaining good validation performance, achieving an mIoU of 67.40\% and an accuracy of 79.25\%. Interestingly, despite having the longest training time of 22.49 seconds per epoch, our proposed EdgeGAT model attained the highest performance metrics among all evaluated architectures, reaching a validation mIoU of 73.44\% and an accuracy of 82.28\%.

\begin{figure}[h]
    \centering
    \includegraphics[width=0.79\linewidth]{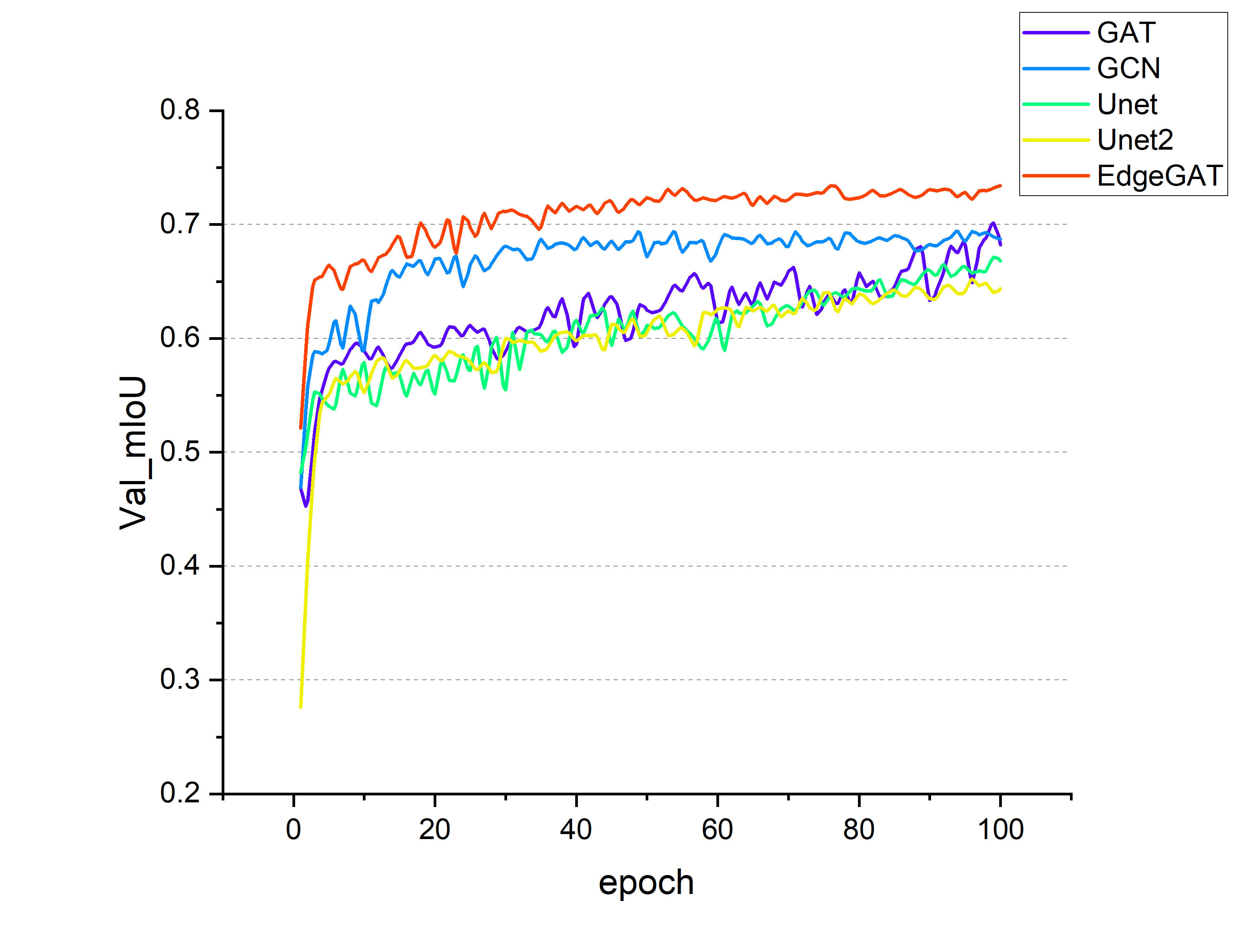}
    \caption{Validation mIoU of GCN, GAT, UNet, UNet2, and EdgeGAT using 13 PCA-extracted features in Ao dataset.
}
    \label{fig:PCAimprovemIoU}
\end{figure}

The effectiveness of incorporating the complete set of 13 PCA-extracted geometric features is further illustrated in Figures~\ref{fig:PCAimproveloss} and~\ref{fig:PCAimprovemIoU}. Figure~\ref{fig:PCAimproveloss} highlights the impact of these features on the loss function during training, clearly showing that EdgeGAT achieved the fastest initial decline, consistently maintaining the lowest loss values throughout training, followed closely by GCN and GAT. Complementarily, Figure~\ref{fig:PCAimprovemIoU} demonstrates a noticeable improvement in validation mIoU across all models when PCA-derived features were included. Notably, EdgeGAT surpassed the 70\% mIoU threshold early on, outperforming traditional models such as GCN and GAT. Moreover, EdgeGAT reached higher mIoU levels as early as epoch 20, further underscoring the effectiveness of combining PCA feature extraction with advanced graph-based learning techniques.

\renewcommand{\arraystretch}{1.2}
\begin{table}[h!]
\centering
\caption{Training Time and Performance Metrics with 13 PCA features on Pheno4D dataset}
\begin{tabular}{lccc}
\toprule
\textbf{Model} & \textbf{Training Time (s)} & \textbf{mIoU (\%)} & \textbf{Val. Accuracy (\%)} \\ 
\midrule
GCN       &  33.60    &  80.28   &  87.80   \\
GAT       &  \textbf{31.30}   &  82.12   &  88.93   \\
UNet      &  82.50   &  77.08  &  85.80  \\
UNet2     &  80.00   &  76.21  &  85.50   \\
EdgeGAT  &  62.80  & \textbf{93.20 }&  \textbf{96.40}  \\
\bottomrule
\end{tabular}
\label{tab:training_time_comparison_allfeat}
\end{table}

\begin{figure}[h!]
    \centering
    \includegraphics[width=0.79\linewidth]{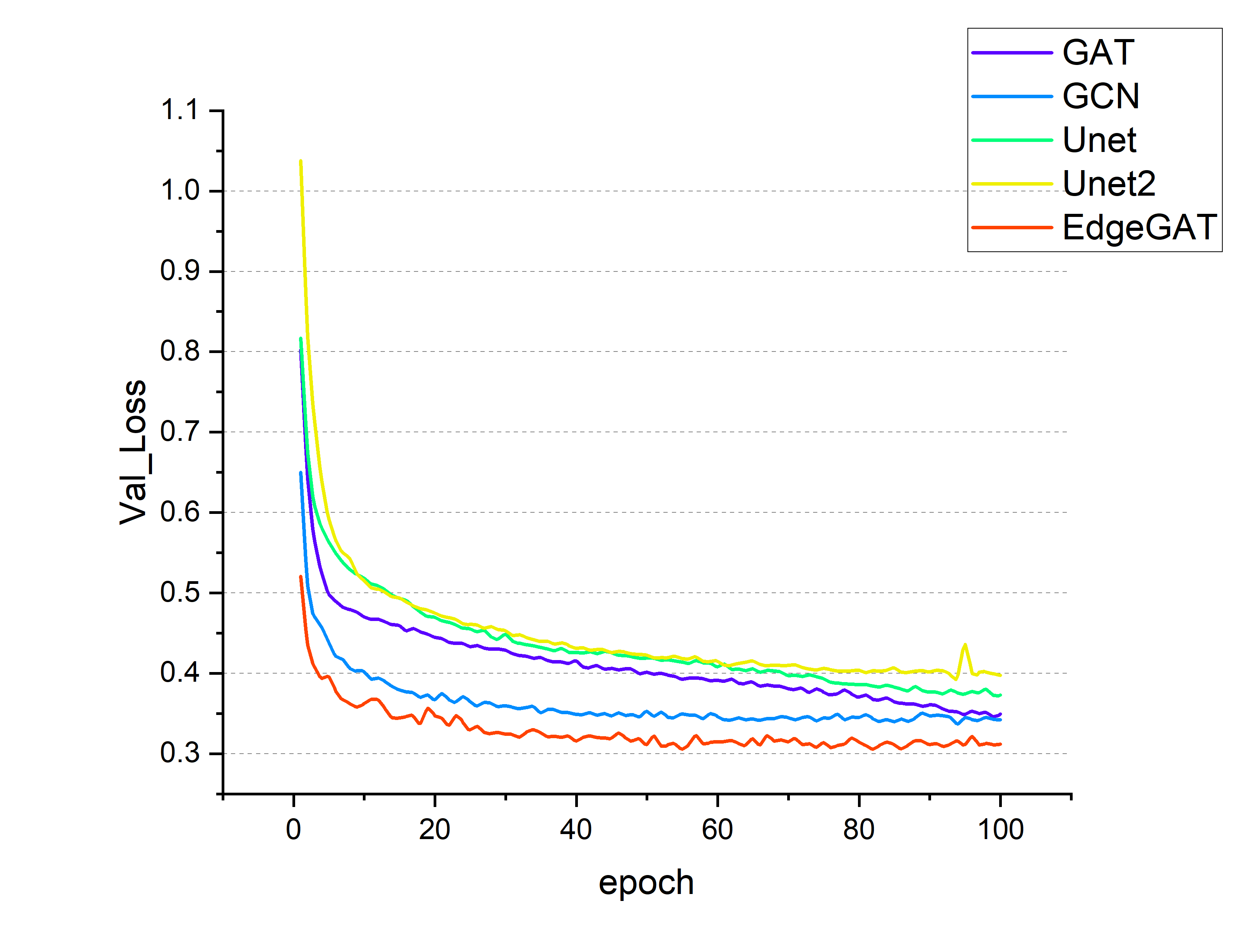}
    \caption{Validation Loss of GCN, GAT, UNet, UNet2, and EdgeGAT using 13 PCA-extracted features.
}
    \label{fig:PCAimproveloss}
\end{figure}  

\begin{figure}[h!]
    \centering
    \includegraphics[width=0.79\linewidth]{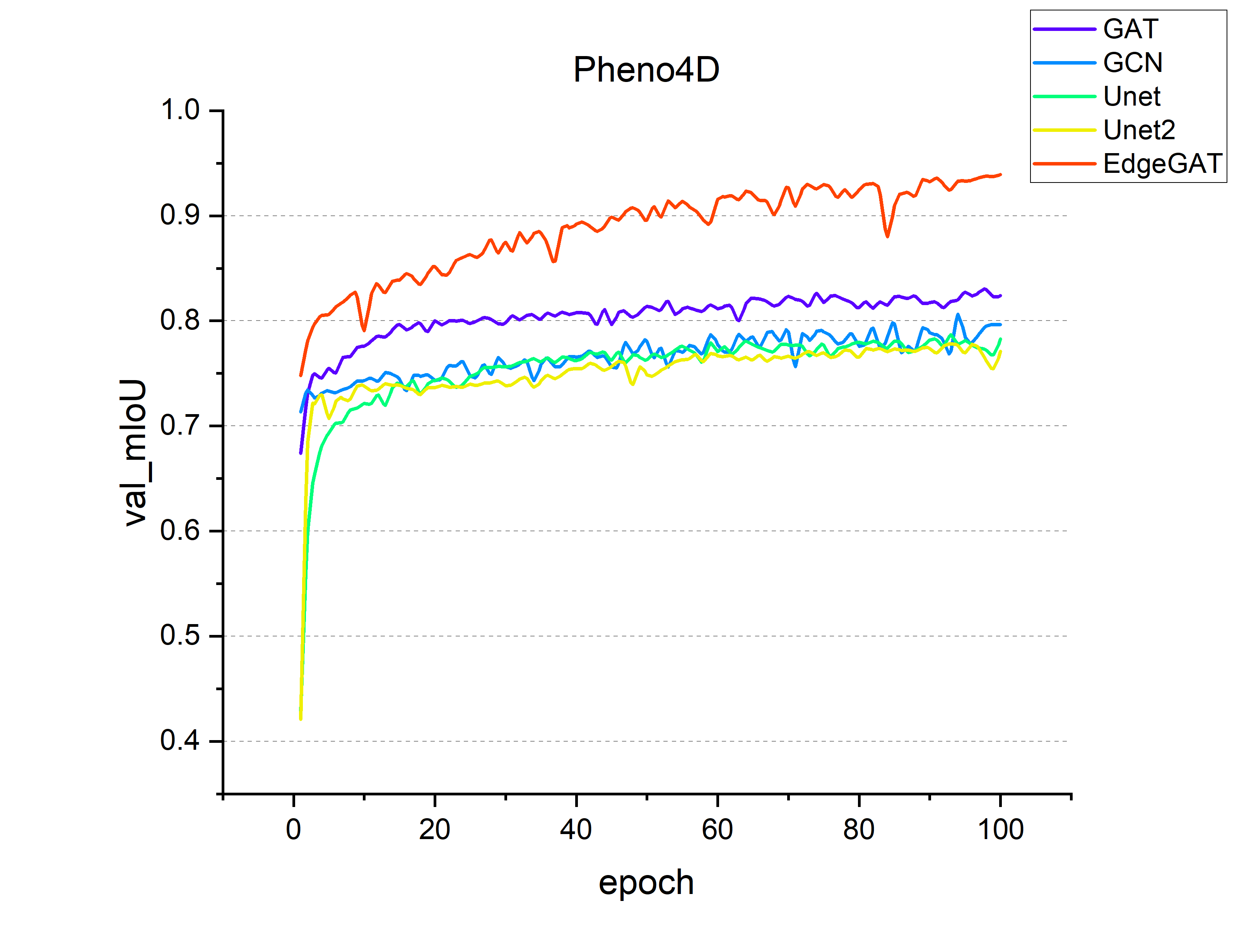}
    \caption{Validation mIoU of GCN, GAT, UNet, UNet2, and EdgeGAT using 13 PCA-extracted features in Pheno4D dataset.
}
    \label{fig:PCAimproveIoUPheno4D}
\end{figure}

Additionally, we assessed the models on the Pheno4D dataset using the full set of 13 PCA-derived features. Table 5 summarises the results. Consistent with previous experiments, the GAT model achieved the shortest training time (31.3 s per epoch), closely followed by the GCN baseline, whereas the U-Net variants remained the slowest. Our proposed EdgeGAT required a moderate 62 s per epoch, yet delivered the best segmentation performance, attaining an mIoU of 93.20 \%. The next-best model was GAT at 82.12 \%, followed by GCN, with both U-Net models trailing behind. Validation accuracy mirrored the mIoU ranking, confirming that the attention-based architectures outperformed their convolution counterparts, with EdgeGAT offering the overall highest accuracy despite its longer training time. Figure \ref{fig:PCAimproveIoUPheno4D} visualises the evolution of the validation mIoU for the same models on the Pheno4D dataset. The curve confirms the numerical trends reported in Table 5: \textbf{EdgeGAT} rises sharply within the first ten epochs and stabilises above the 90 \% mark, surpassing the other graph-based architectures by nearly ten percentage points. The standard GAT model follows with values consistently above 80 \%, while the GCN baseline converges just below that level. Both U-Net variants remain well behind, never exceeding the 70 \% threshold. The early and sustained lead of EdgeGAT highlights not only its higher final accuracy but also its faster convergence and greater stability throughout training.

This trade-off between computational efficiency and predictive performance highlights the practical considerations when selecting an architecture. While GCN and GAT offer excellent speed–performance balance, models like EdgeGAT are better suited when maximum mIoU is critical, even at the expense of longer training durations.

\subsection{Comparative Analysis and Advantages of EdgeGAT over other Point Cloud Models}

In this section, we conduct additional comparative experiments involving both point-based (PointNet and DGCNN) and graph-based (GAT and EdgeGAT) architectures, aiming to evaluate the relative effectiveness of our proposed EdgeGAT model. The analysis specifically focuses on the validation loss and mean Intersection over Union metrics.

\begin{figure}[h!]
    \centering
    \includegraphics[width=0.79\linewidth]{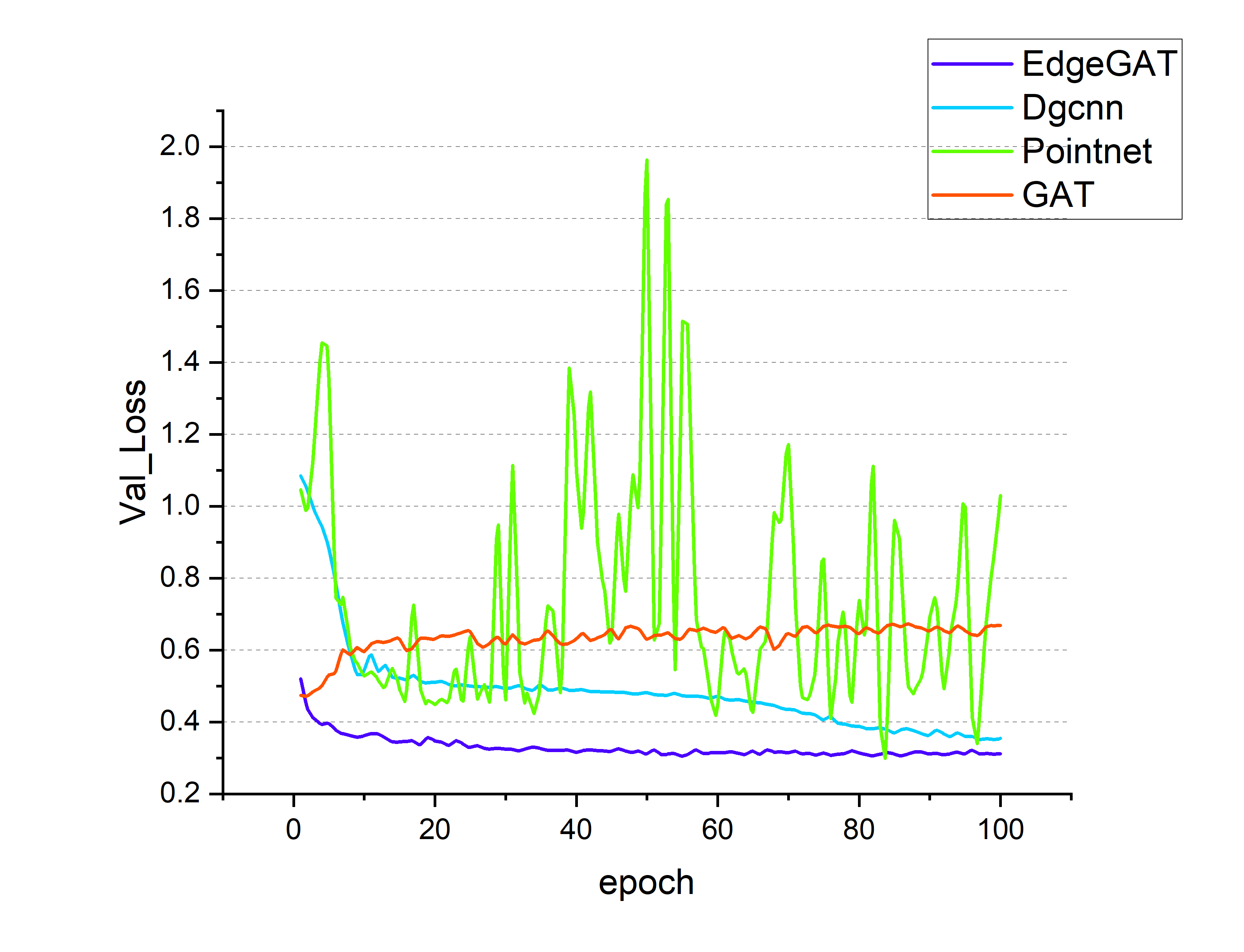}
    \caption{Validation Loss of point and graph-based models using 13-PCA features: EdgeGAt, DGCNN, Pointnet and GAT.}
    \label{fig:LosscomparePointModelsa}
\end{figure}

Figure~\ref{fig:LosscomparePointModelsa} illustrates the validation loss across epochs for each architecture. The EdgeGAT model demonstrates superior convergence, achieving the lowest loss values rapidly after approximately epoch 30. In contrast, PointNet exhibits high variability without reaching a stable equilibrium, reflecting its difficulty in consistently modeling the three classes. The DGCNN model shows a steady but slower decrease in validation loss, approaching values similar to EdgeGAT only after epoch 80. Interestingly, GAT experiences signs of potential overfitting or training instability, as its loss initially starts at around 0.50 but gradually increases to approximately 0.60, highlighting potential challenges in its convergence behavior when using the complete set of PCA-derived features.

\begin{figure}[h]
    \centering
    \includegraphics[width=0.79\linewidth]{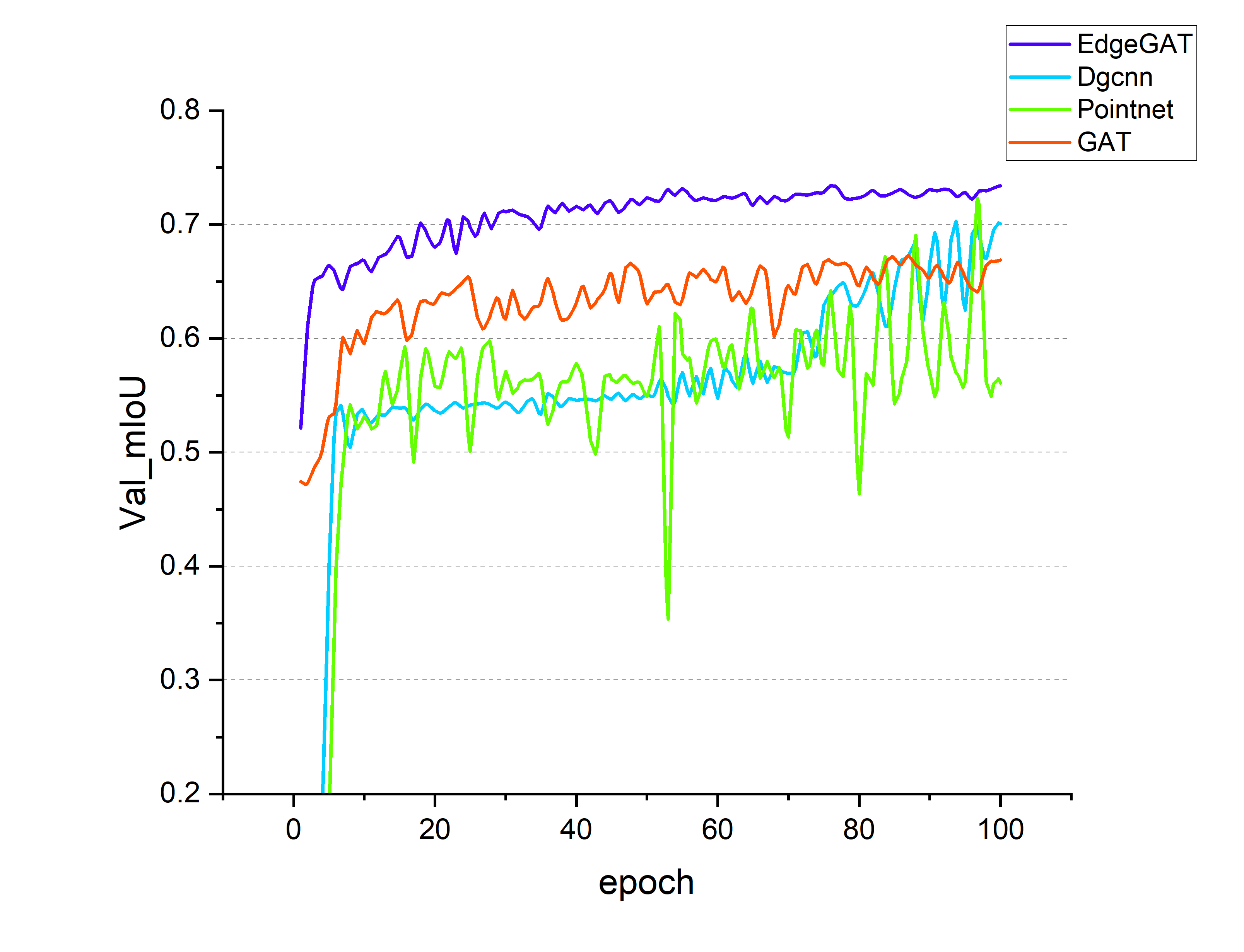}
    \caption{Validation mIoU of point and graph-based models using 13-PCA features: EdgeGAT, DGCNN, Pointnet and GAT.}
    \label{fig:IoUcomparePointModelsb}
\end{figure}

The validation mIoU for each architecture is presented in Figure~\ref{fig:IoUcomparePointModelsb}. Here, our EdgeGAT model achieves the highest and most consistent mIoU performance, surpassing 73\%. DGCNN follows, reaching a maximum near 70\%, but only towards the final epochs of training. GAT remains consistently below 70\%, aligning with its observed loss instability. PointNet occasionally reaches peaks around 70\%, yet its performance fluctuates significantly, ultimately averaging around 61\%, underscoring limited consistency in capturing class-specific features.

\renewcommand{\arraystretch}{1.2}
\begin{table}[h!]
\centering
\caption{Performance Comparison of Evaluated Architectures in Ao Dataset}
\label{tab:validation_comparison_models}
\begin{tabular}{lcccc}
\toprule
\textbf{Metric} & \textbf{EdgeGAT} & \textbf{DGCNN} & \textbf{PointNet} & \textbf{GAT} \\ 
\midrule
Loss (\%)        & \textbf{29.28} & 31.71 & 43.13 & 32.85 \\[2pt]
mIoU (\%)        & \textbf{73.35} & 70.54 & 60.07 & 70.31 \\[2pt]
Accuracy (\%)    & \textbf{85.09} & 83.53 & 77.53 & 83.05 \\[2pt]
Precision (\%)   & 82.89 & \textbf{83.14} & 55.82 & 81.21 \\
T. Time (s) & 22.49 & 189.29 & 16.20 & \textbf{11.29} \\
\bottomrule
\end{tabular}
\end{table}

In addition to the results illustrated previously, Table~\ref{tab:validation_comparison_models} summarizes a detailed quantitative comparison across all evaluated architectures on the validation dataset. Specifically, it includes performance metrics such as loss, mean IoU, accuracy, precision, and training time per epoch. While DGCNN achieved the highest precision of 83.14\%, it required the longest training duration of approximately 189 seconds per epoch. Conversely, the GAT model recorded the shortest training time at 11.09 seconds per epoch; however, it did not reach the performance obtained by our proposed EdgeGAT model. Our EdgeGAT architecture, despite having a moderate training time of 22.30 seconds per epoch, outperformed other models by achieving the highest values for mIoU and accuracy, along with the lowest loss values. Lastly, the PointNet model demonstrated a relatively quick training time of 16.20 seconds per epoch but exhibited the lowest performance metrics among all evaluated architectures.

\begin{figure}[h!]
\centering
\includegraphics[width=0.79\linewidth]{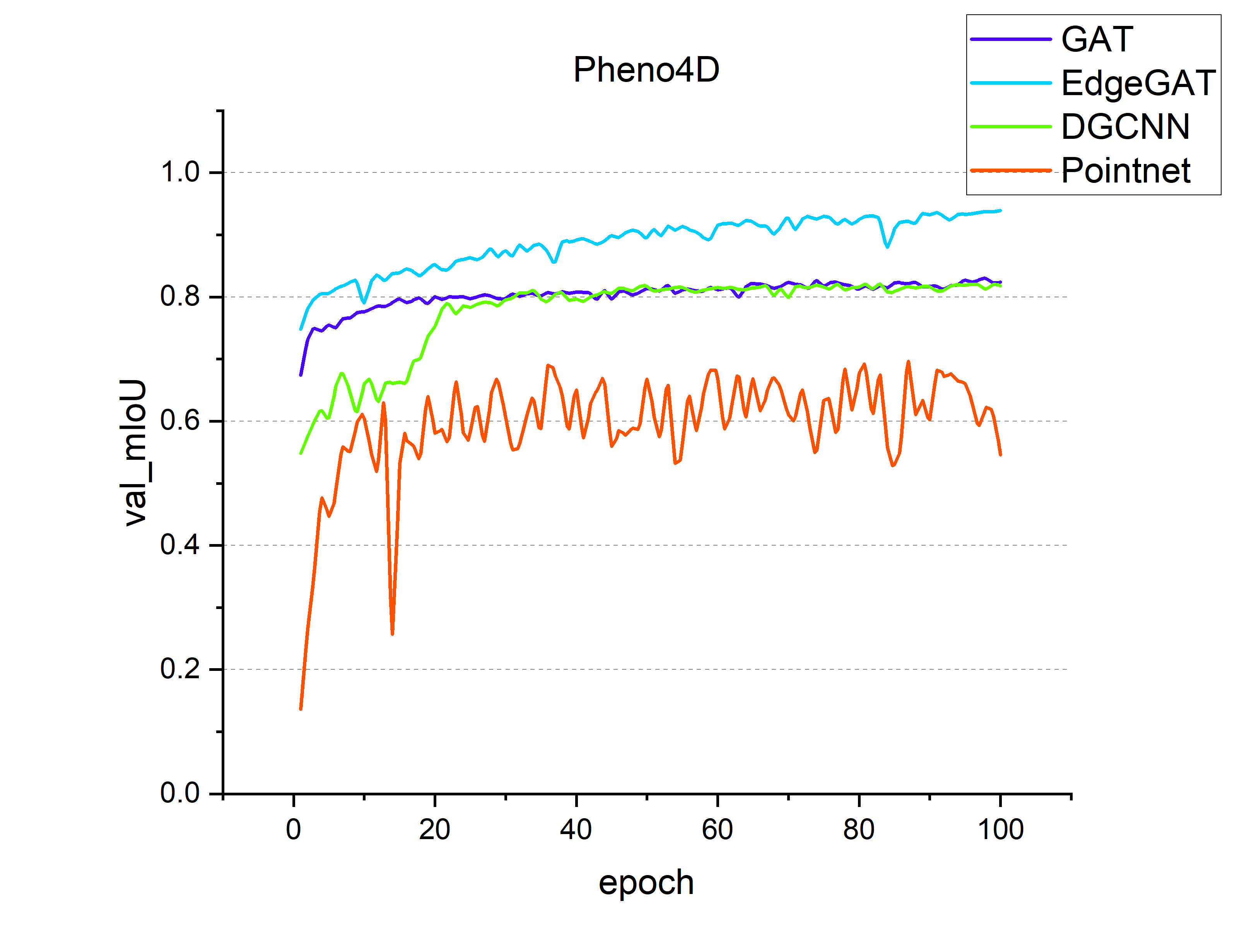}
\caption{Validation mIoU curves for EdgeGAT, DGCNN, PointNet, and GAT on the Pheno4D dataset.}
\label{fig:pheno4d_miou_curve}
\end{figure}

\medskip
\noindent\textbf{Evaluation on the Pheno4D Dataset.}
Table~\ref{tab:validation_comparison_pheno4d} summarizes the performance of all evaluated architectures on the more complex and diverse Pheno4D dataset, confirming the robustness and generalizability of our EdgeGAT model. Consistent with earlier results from the Ao dataset, EdgeGAT significantly surpasses the other architectures, achieving a notably high mIoU of 93.20\%. This result outperforms the next-best models, GAT and DGCNN, by approximately 11 percentage points, as both models converge to values slightly above 80\%. In contrast, PointNet exhibits considerably lower performance, achieving an mIoU of only around 60\%, highlighting its limited capacity for consistently capturing complex geometric features present in this dataset. These trends are clearly reflected in Figure~\ref{fig:pheno4d_miou_curve}, where the validation mIoU curves illustrate that EdgeGAT not only reaches superior accuracy levels, but also attains this performance much earlier during training, emphasizing its stability and efficiency in learning rich feature representations from challenging real-world point cloud data.

\renewcommand{\arraystretch}{1.2}
\begin{table}[h!]
\centering
\caption{Performance Comparison of Evaluated Architectures in Pheno4D Dataset}
\label{tab:validation_comparison_pheno4d}
\begin{tabular}{lcccc}
\toprule
\textbf{Metric} & \textbf{EdgeGAT} & \textbf{DGCNN} & \textbf{PointNet} & \textbf{GAT} \\ 
\midrule
Loss (\%)        & \textbf{04.09} & 12.29 & 24.96 & 16.94 \\[2pt]
mIoU (\%)        & \textbf{93.20} & 82.05 & 60.82 & 82.12 \\[2pt]
Accuracy (\%)    & \textbf{96.40} & 95.43 & 86.28 & 88.93 \\[2pt]
Precision (\%)   & \textbf{97.20} & 89.12 & 80.25 & 89.46 \\
T. Time (s) & 62.80 & 452.17 & 36.00 & \textbf{31.30} \\
\bottomrule
\end{tabular}
\end{table}

Collectively, these results emphasize that our proposed EdgeGAT model not only reaches superior mIoU and validation accuracy values, but also converges faster and more reliably compared to both point-based and alternative graph-based approaches when integrating the full set of PCA-extracted features.

\begin{figure}[h!]
    \centering
    \includegraphics[width=0.65\linewidth]{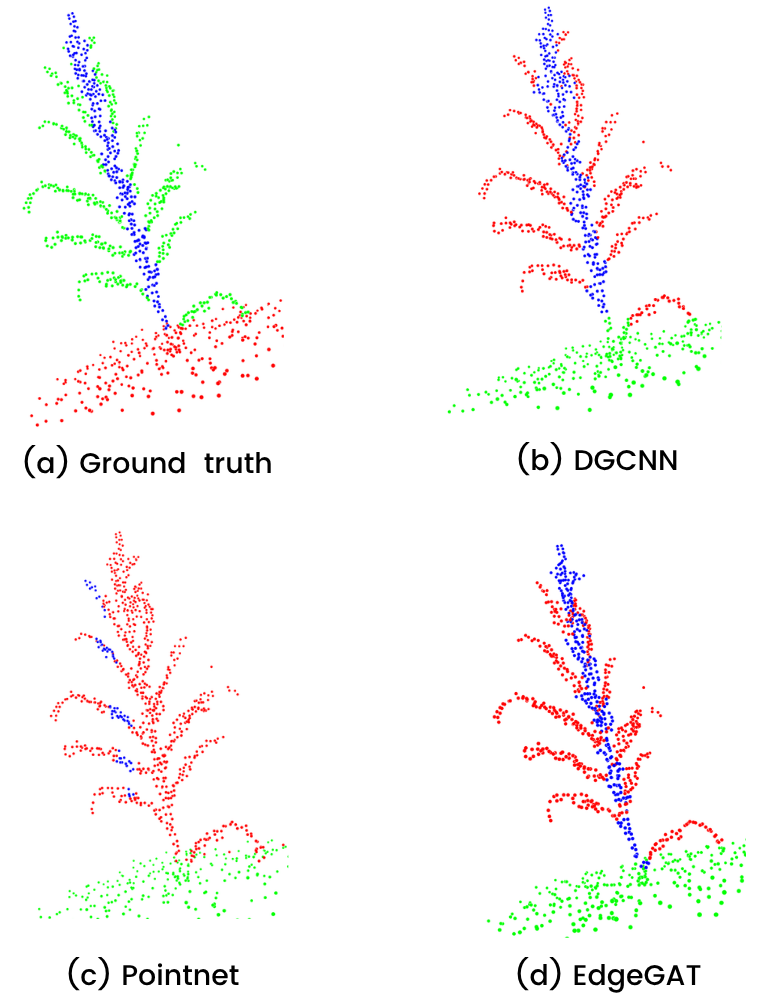}
    \caption{Visualization of point cloud classification: (a) Ground truth of maize plant, (b) Predicted with DGCNN model, (c) Predicted with pointnet model, (d) Predicted with our EdgeGAT model.}
    \label{fig:corncomparation}
\end{figure}

In order to better visualize the performance of each model in a real-world scenario, we selected a raw 3D point cloud from the test set and evaluated it using the best checkpoint from each model. These results are shown in Figure~\ref{fig:corncomparation}. Part (a) presents the original input point cloud. Part (b) displays the prediction output from DGCNN, while part (c) corresponds to the result obtained with PointNet. Finally, part (d) shows the prediction from our EdgeGAT model. As observed, PointNet only predicted two out of the three trained classes, demonstrating limited expressiveness. DGCNN performed noticeably better, closely matching the output of our model. However, EdgeGAT exhibited more accurate class separation and clearer segmentation boundaries.

\section{Discussion}
In recent years, methods for point cloud segmentation have significantly evolved, transitioning from traditional convolutional neural network approaches, often hindered by the irregularity and unordered nature of point cloud data, to graph-based methods that explicitly exploit the relational structure inherent to 3D information. Graph-based models represent point clouds as structured graphs, treating each point as a vertex interconnected by edges that capture local geometric relationships. By aggregating local context through these edges, these models facilitate precise feature extraction, crucial for effectively segmenting complex structures such as maize plants.

The integration of attention mechanisms within graph-based models further refines segmentation performance by adaptively weighting the contribution of neighboring vertices. Such attention-based modules allow the model to dynamically prioritize relevant local features, addressing challenges related to uneven feature representation and local ambiguities. In this context, our proposed EdgeGAT model clearly demonstrates these advantages. By combining edge convolutional operations enhanced with Residual MLPs and attention-based mechanisms from Graph Attention Networks, the EdgeGAT architecture achieves superior feature representation, faster convergence, and significantly improved segmentation accuracy compared to both standard GAT and alternative point-based models.

Our extensive evaluations across two different datasets (Ao and Pheno4D) further reinforced these observations. On the Ao dataset, EdgeGAT outperformed alternative architectures, achieving higher and more consistent mIoU and accuracy metrics. Interestingly, evaluations on the Pheno4D dataset, which consists of younger maize plants approximately ten days old, yielded even higher performance metrics due to the reduced structural complexity of the plants at this developmental stage. Specifically, EdgeGAT attained an mIoU exceeding 90\%, substantially outperforming DGCNN and GAT, both of which reached approximately 80\%. PointNet consistently lagged behind, underscoring its limitations in reliably capturing geometric features even in less structurally complex scenarios. Despite EdgeGAT relatively higher computational demands compared to simpler models, the substantial gains in accuracy, stability, and rapid convergence strongly justify its adoption, particularly when precise segmentation and robust performance across different developmental stages of plants are critical.

\section{Conclusion}

In this work, we demonstrated the effectiveness of our proposed model, EdgeGAT, for accurate segmentation and classification of maize plant components. By enriching the original 3D coordinates through PCA-based geometric feature extraction, we significantly improved the representation of local geometric details, enabling the construction of more informative graphs and facilitating accurate differentiation among plant structures.

Our proposed EdgeGAT architecture, which integrates EdgeConv modules enhanced with Residual MLPs and Graph Attention layers, proved highly effective in capturing both fine-grained local characteristics and broader contextual relationships within the point clouds. Specifically, by leveraging edge-based convolutions and dynamic neighbor aggregation, EdgeGAT achieved superior performance, exhibiting notably faster convergence and higher segmentation accuracy compared to traditional graph-based methods and other state-of-the-art point cloud architectures.

However, our EdgeGAT model also presents certain limitations. Although it achieves strong accuracy and precision in segmentation tasks, even with limited samples, it may not generalize effectively to other tasks, such as classification of 2D images or datasets in which the axis alignment among classes is less consistent than in structured scenarios like plant segmentation. Future work should therefore explore the model generalizability by evaluating its performance on diverse point cloud datasets involving semantic segmentation in different contexts and applications.

Overall, the progression toward graph-based segmentation models, particularly those leveraging edge-based convolution and attention mechanisms, presents significant potential for improving the accuracy and efficiency of point cloud segmentation. This paradigm is particularly relevant in agricultural applications, where precise identification and segmentation of plant components—such as stems, leaves, and soil regions—are crucial for precision agriculture tasks, including crop monitoring, biomass estimation, and targeted agricultural interventions.

\section*{Acknowledgements}

The authors wish to express their sincere gratitude to CINVESTAV for their institutional support throughout this project. We also thank CONAHCYT for the financial assistance provided, which was instrumental in enabling this research. Additionally, we extend our appreciation to the authors of the original datasets, whose efforts in data collection and sharing made this work possible.

\appendix
\section{Appendix A: Evaluated Sub-architectures}
\label{app1}

This appendix provides detailed descriptions of two alternative GCN-based sub-architectures evaluated during preliminary experiments: a Simple GCN model and a Graph U-Net architecture.
\subsection*{A.1 Simple Graph Convolutional Network (Simple GCN)}

The Simple GCN model consists of multiple stacked GCN layers (\texttt{GCNConv}), each followed by a ReLU activation and dropout regularization, except for the final output layer, which outputs raw logits directly. 

Mathematically, a single GCN convolutional layer can be described as follows:

\begin{equation}
    \mathbf{H}^{(l+1)} = \sigma\left(\hat{\mathbf{D}}^{-\frac{1}{2}}\hat{\mathbf{A}}\hat{\mathbf{D}}^{-\frac{1}{2}}\mathbf{H}^{(l)}\mathbf{W}^{(l)}\right),
\end{equation}

where \(\mathbf{H}^{(l)}\) represents vertex embeddings at layer \(l\), \(\mathbf{W}^{(l)}\) are learnable weight matrices, \(\hat{\mathbf{A}}=\mathbf{A}+\mathbf{I}\) is the adjacency matrix of the graph with added self-connections, and \(\hat{\mathbf{D}}\) is the diagonal degree matrix of \(\hat{\mathbf{A}}\). The activation function \(\sigma(\cdot)\) is ReLU in intermediate layers.

The final classification logits \(\mathbf{Z}\) for vertices are computed directly from the last GCN layer without activation:

\begin{equation}
    \mathbf{Z} = \hat{\mathbf{D}}^{-\frac{1}{2}}\hat{\mathbf{A}}\hat{\mathbf{D}}^{-\frac{1}{2}}\mathbf{H}^{(L-1)}\mathbf{W}^{(L-1)},
\end{equation}

where \(L\) is the total number of layers.

\textbf{GCN Model:}  
The diagram should depict a clear stack of multiple GCNConv layers, each followed by ReLU and dropout, showing explicitly the skip of activation in the last layer.

\begin{figure}[h!]
    \centering
    \includegraphics[width=0.85\linewidth]{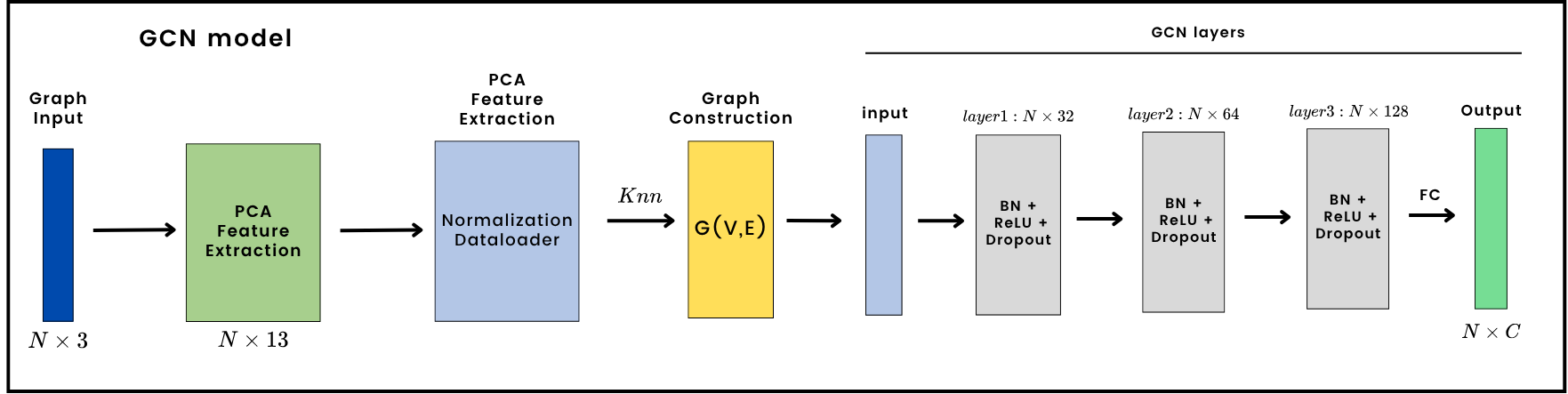}
    \caption{Visualization of GCN model utilized in ablation studies}
    \label{fig:GCNarchitecture}
\end{figure}

\subsection*{A.2 Graph U-Net with GCN (GCN-UNet)}

The second architecture is a Graph U-Net, which integrates pooling and unpooling layers with GCN convolutions. It is inspired by the U-Net structure commonly used in computer vision, enabling the network to capture hierarchical features at multiple scales within graph data.

\textbf{GCN Unet Model:}  
The diagram should depict a clear stack of multiple GCNConv layers, each followed by ReLU and dropout, showing explicitly the skip of activation in the last layer.

\begin{figure}[h!]
    \centering
    \includegraphics[width=0.85\linewidth]{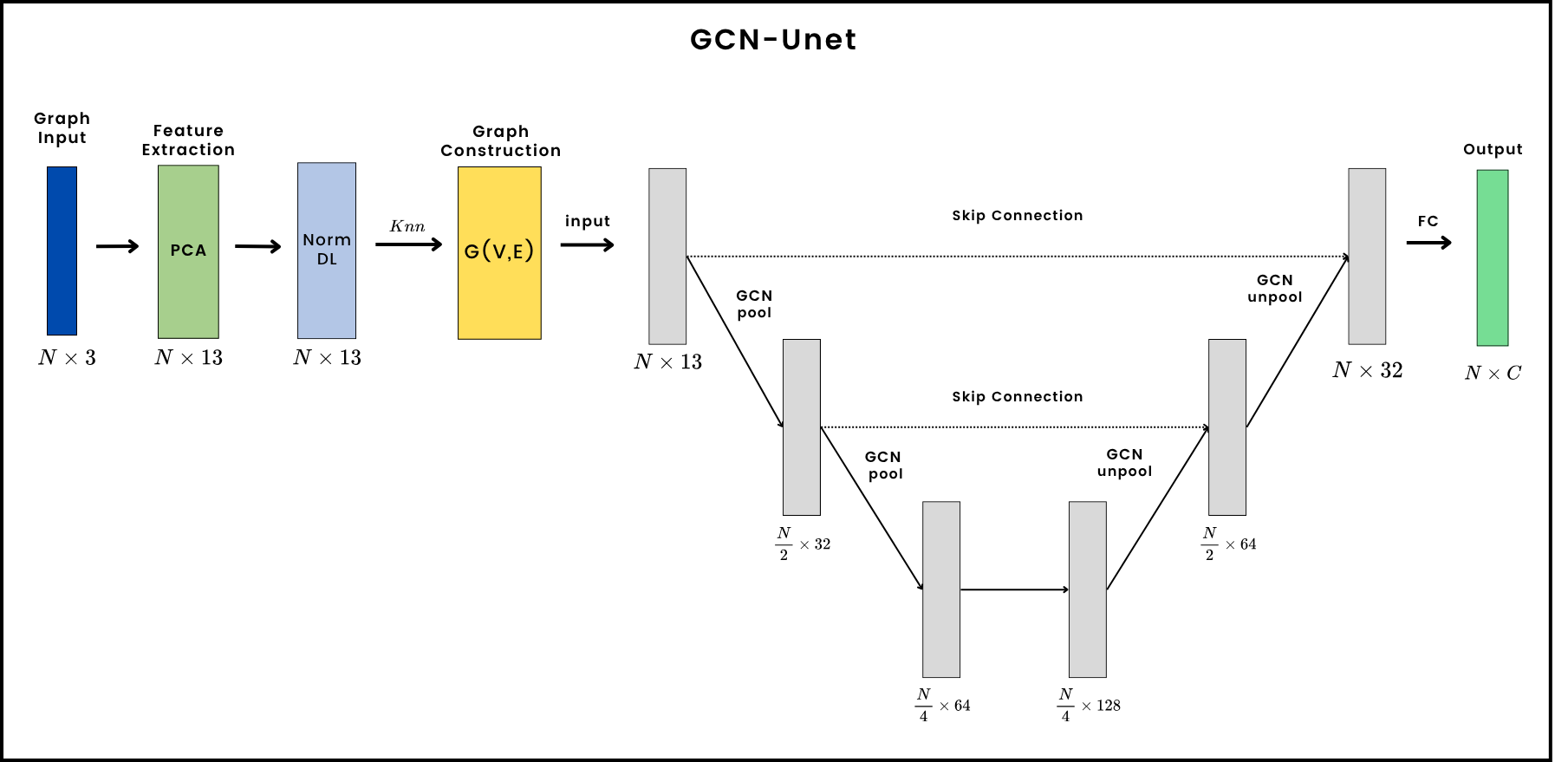}
    \caption{Visualization of GCN-Unet model utilized in ablation studies}
    \label{fig:Unetarchitecture}
\end{figure}

The Graph U-Net operates through three main steps:

\begin{enumerate}
    \item \textbf{Encoder:} Repeated application of GCN layers and graph pooling to reduce the number of vertices progressively.
    \item \textbf{Bottleneck:} A latent representation is obtained at the deepest level.
    \item \textbf{Decoder:} Gradual reconstruction of the original graph resolution by graph unpooling and additional GCN layers, integrating skip connections from the encoder stage.
\end{enumerate}

Formally, at each encoding and decoding step, the vertex embeddings are computed as:

\begin{equation}
    \mathbf{H}^{(l+1)} = \text{GCN}\left(\mathbf{H}^{(l)}, \mathbf{A}^{(l)}\right),
\end{equation}

where \(\mathbf{H}^{(l)}\) is the vertex embedding at layer \(l\), and \(\mathbf{A}^{(l)}\) denotes the adjacency matrix at layer \(l\), updated via pooling/unpooling operations.

The final vertex representation, incorporating skip connections, is given by:

\begin{equation}
    \mathbf{H}^{(\text{final})} = \mathbf{H}^{(L)} + \sum_{l=1}^{L-1}\mathbf{H}^{(l)},
\end{equation}

where \(\mathbf{H}^{(L)}\) is the output of the decoder, and the summation represents the residual connections from the encoder.

\subsection*{A.3 Enhanced Graph U-Net2}

We also evaluated an enhanced version of the Graph U-Net, referred to as \textbf{GCN-UNet2}, which closely follows the original architecture described previously but incorporates two significant additions aimed at improving stability and regularization. Specifically, after each GCN convolutional layer within both encoder and decoder stages, a \textbf{Batch Normalization} operation followed by \textbf{Dropout} (
\( p = 0.2 \) ) was applied. This modification ensures more stable training dynamics, reduces potential overfitting, and helps maintain consistent gradient flow throughout the network.

Formally, the vertex embeddings after each enhanced convolutional layer are computed as:

\begin{equation}
\mathbf{H}^{(l+1)} = \text{Dropout}\left(\text{BatchNorm}\left(\text{GCN}\left(\mathbf{H}^{(l)}, \mathbf{A}^{(l)}\right)\right)\right),
\end{equation}

\noindent where $H^{(l)}$ denotes the vertex embedding at layer$l$, and $A^{(l)}$ is the corresponding adjacency matrix. All other structural components, including the encoder-decoder arrangement, pooling, unpooling, and skip connections, remain identical to the original Graph U-Net implementation.

\subsection*{A.4 Graph Attention Network (}

The GAT architecture uses the attention mechanisms to dynamically weigh the influence of neighboring vertices. In contrast to standard graph convolutional networks, GAT employs a self-attention mechanism to assign adaptive importance scores to each neighbor during aggregation, allowing it to effectively capture varying degrees of local relevance in the input graph.

The GAT architecture comprises two key steps at each convolutional layer:

\begin{enumerate}
    \item \textbf{Feature Transformation:} The initial vertex features \(\mathbf{h}_i\) are linearly projected into a higher-dimensional space through learnable weight matrices.
    \item \textbf{Attention-based Aggregation:} Attention coefficients \(\alpha_{ij}\) are computed for each pair of connected vertices \((i,j)\), determining the contribution of vertex \(j\)'s transformed features to the updated representation of vertex \(i\). Formally, the attention coefficients are computed as:
\end{enumerate}

\begin{equation}
    \alpha_{ij} = \frac{\exp\left(\text{LeakyReLU}\left(\mathbf{a}^{\top}[\mathbf{W}\mathbf{h}_i \| \mathbf{W}\mathbf{h}_j]\right)\right)}{\sum_{k \in \mathcal{N}(i)} \exp\left(\text{LeakyReLU}\left(\mathbf{a}^{\top}[\mathbf{W}\mathbf{h}_i \| \mathbf{W}\mathbf{h}_k]\right)\right)},
\end{equation}

\noindent where \(\mathbf{a}\) is a learnable attention weight vector, \(\mathbf{W}\) represents the linear transformation matrix, and \(\mathcal{N}(i)\) denotes the set of neighboring vertices \(i\).

Finally, the updated vertex embedding is computed by aggregating these features weighted by their attention scores:

\begin{equation}
    \mathbf{h}'_i = \sigma\left(\sum_{j \in \mathcal{N}(i)} \alpha_{ij}\mathbf{W}\mathbf{h}_j\right),
\end{equation}

\noindent where \(\sigma(\cdot)\) denotes a nonlinear activation function, typically a LeakyReLU.

\begin{figure}[h!]
    \centering
    \includegraphics[width=0.85\linewidth]{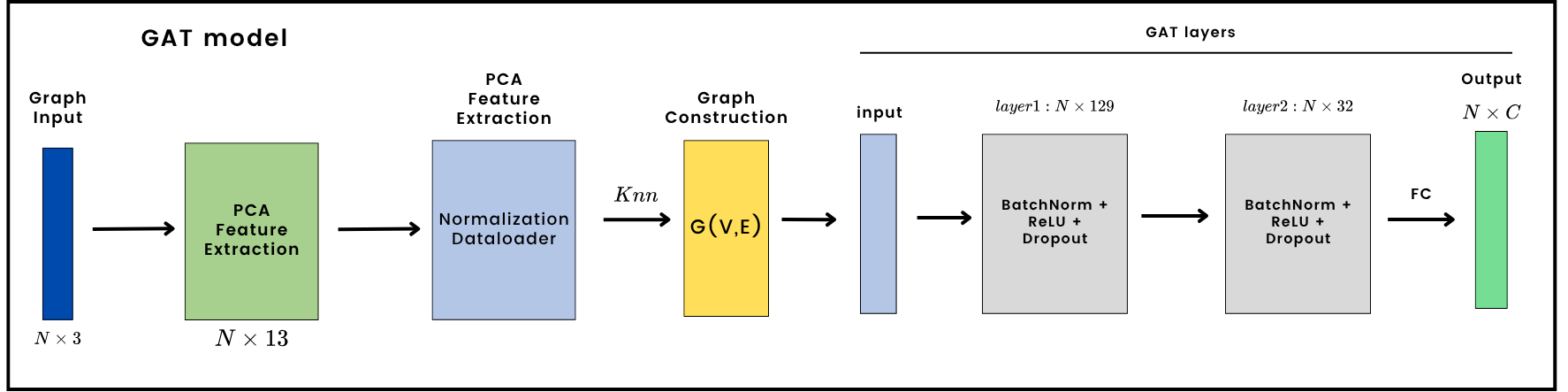}
    \caption{Visualization of the Graph Attention Network (GAT) architecture employed as a baseline in the comparative studies.}
    \label{fig:GATarchitecture}
\end{figure}

The complete GAT architecture used in our experiments employs two consecutive GAT convolutional layers, each followed by batch normalization, nonlinear activation, and dropout regularization, as illustrated in Figure~\ref{fig:GATarchitecture}.

\subsection*{Additional Notes}

The architectures described in this appendix served as baseline models to benchmark the performance of more advanced graph-based approaches, particularly EdgeConv and GAT-based networks. Results obtained from these simpler models provided valuable insights into the effectiveness of incorporating residual connections, multi-scale graph operations, and attention mechanisms, which ultimately guided the design and refinement of our proposed EdgeGAT architecture.



\bibliographystyle{elsarticle-num} 
\bibliography{mybibfile}

\end{document}